\documentclass[journal]{IEEEtran}
\usepackage{amsmath,amsfonts}
\usepackage{algorithm}
\usepackage{array}
\usepackage[caption=false,font=normalsize,labelfont=sf,textfont=sf]{subfig}
\usepackage{textcomp}
\usepackage{stfloats}
\usepackage{url}
\usepackage{verbatim}
\usepackage{graphicx}
\graphicspath{{figures/}}
\usepackage{cite}
\usepackage{amssymb}
\usepackage{multirow}
\usepackage{xcolor}
\usepackage{bm}
\usepackage{enumerate}
\usepackage{booktabs}
\usepackage{hyperref}
\DeclareMathOperator*{\argmin}{argmin}
\usepackage{siunitx}

\usepackage{algpseudocode}
\usepackage{tablefootnote}
\usepackage{amssymb}
\usepackage{color, colortbl}
\usepackage{pifont}%
\usepackage{amsmath}
\newcommand{\cmark}{\ding{51}}%
\newcommand{\xmark}{\ding{55}}%
\definecolor{pink}{rgb}{0.858, 0.188, 0.478}

\newcommand{\minisection}[1]{\vspace{0.04in} \noindent {\bf #1}\ \ }
\usepackage{breqn}
\definecolor{LightCyan}{rgb}{0.5647, 0.5647, 0.5647}

\usepackage{soul}
\setul{0.5ex}{0.2ex}

\begin{document}
\author{
Marco Cotogni$^{1,2,*}$, Jacopo Bonato$^{1,*}$, Luigi Sabetta$^{1,*}$, Francesco Pelosin$^{3,\dag}$, Alessandro Nicolosi$^{1}$\\ \vspace{2.5mm}
{$^{1}$Leonardo Labs,  $^{2}$University of Pavia, $^{3}$Covision Lab}
\\\vspace{2.5mm}
\tt\small{\{marco.cotogni, jacopo.bonato, luigi.sabetta\}.ext@leonardo.com}
\tt\small{francesco.pelosin@covisionlab.com, alessandro.nicolosi@leonardo.com}
\thanks{$^*$Equal Contributions\\ $\dag$Work Performed at Leonardo Labs, now at Covision Lab}
}

\title{DUCK: Distance-based Unlearning via Centroid Kinematics}

\maketitle

\begin{abstract}
Machine Unlearning is rising as a new field, driven by the pressing necessity of ensuring privacy in modern artificial intelligence models. This technique primarily aims to eradicate any residual influence of a specific subset of data from the knowledge acquired by a neural model during its training. This work introduces a novel unlearning algorithm, denoted as \textbf{Distance-based Unlearning via Centroid Kinematics (DUCK)}, which employs metric learning to guide the removal of samples matching the nearest incorrect centroid in the embedding space. Evaluation of the algorithm's performance is conducted across various benchmark datasets in two distinct scenarios, class removal, and homogeneous sampling removal, obtaining state-of-the-art performance.
We also introduce a novel metric, called Adaptive Unlearning Score (AUS), encompassing not only the efficacy of the unlearning process in forgetting target data but also quantifying the performance loss relative to the original model.
Additionally, we conducted a thorough investigation of the unlearning mechanism in DUCK, examining its impact on the organization of the feature space and employing explainable AI techniques for deeper insights.

\noindent Code will be published upon acceptance \footnote{This work has been submitted to the IEEE for possible publication. Copyright may be transferred without notice, after which this version may no longer be accessible.}.
\end{abstract}


\begin{figure*}[!ht]
    \centering
    \includegraphics[width=.9\linewidth]{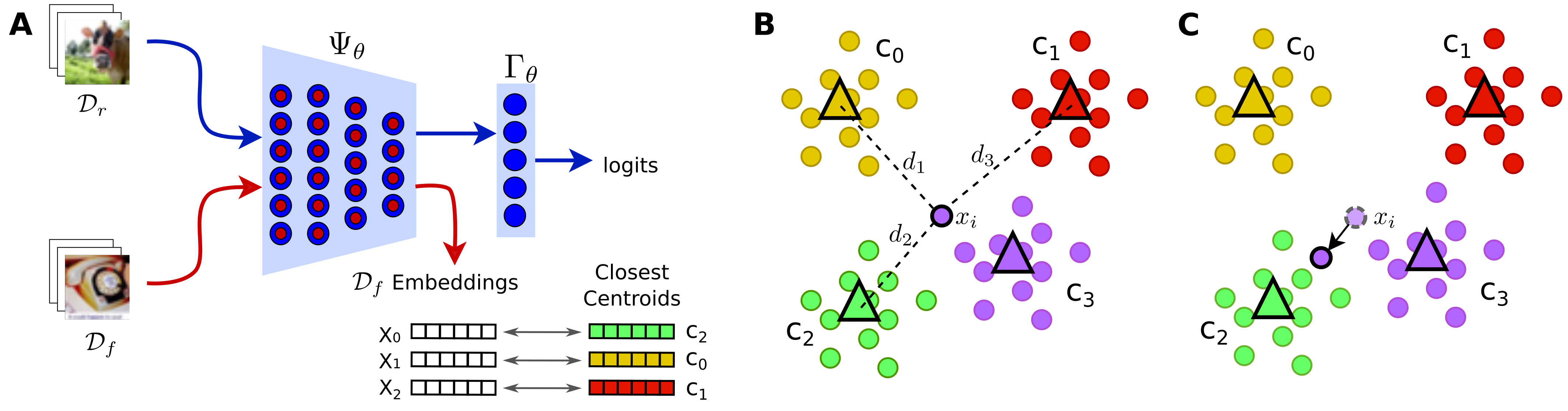}
    \caption{DUCK unlearning scheme. A) DUCK architecture $\Phi_{\theta} = \Gamma_{\theta} \circ \Psi_{\theta}$ during unlearnig phase. The weights of $\Gamma_{\theta}$ are updated through gradient descent using $\mathcal{L}_{RET}$ loss on $\mathcal{D}_r$. Importantly, weights of $\Psi_{\theta}$ are optimized using both the gradient obtained from $\Gamma_{\theta}$ and also from the closest-centroid matching applied on $\mathcal{D}_f$. B) Representation of closest-centroid matching procedure. For each sample $x_i \in \mathcal{D}_f$ its distance from incorrect classes centroids is computed and the closest one is selected. C) The distance between $x_i$ and the selected closest centroid is minimized through gradient descent.}
    \label{fig:scheme}
\end{figure*}
\section{Introduction}
\label{sec:intro}

In the ever-evolving landscape of artificial intelligence and machine learning, the development of algorithms capable of learning and adapting to data has garnered significant attention. The ascent of artificial intelligence algorithms enabled remarkable achievements in various domains. Nevertheless, the use of machine learning systems comes with some privacy challenges. Machine learning models, once trained, store in their parameters the patterns and information extracted from the data they have been exposed to.
Unfortunately, this persistence of learned information can lead to unintended consequences, from privacy breaches due to lingering personal data, to biased decisions rooted in historical biases in the training data.  

For instance, the information extraction can be performed by a malicious attacker through several procedures \cite{mia1, mia2,mia3,mia4, nguyen2022survey, hu2022membership}. Similarly, the injection of adversarial data can force machine learning models to output wrong predictions resulting in serious security problems \cite{ren2020adversarial,cao2015towards,marchant2022hard}. Last but not least, numerous recent privacy-preserving regulations, have been enacted, such as the European Union's General Data Protection Regulation\cite{magdziarczyk2019right} and the California Consumer Privacy Act \cite{pardau2018california}. Under these regulations, users are entitled to request the deletion of their data and related information to safeguard their privacy. Hence, a procedure to remove selectively part of the information stored in the parameters of the models represents a fundamental open problem. 

To this end, Machine Unlearning studies methodologies to instill selective forgetting in trained models such that training data or sensitive information can not be recovered \cite{nguyen2022survey,xu2023machine,shaik2023exploring, mercuri2022introduction, villaronga2018humans}. In the context of an original model trained on a dataset, unlearning approaches seek to eliminate information about a specific subset of data, termed the forget-set, from the parameters of the original network while preserving its generalization performance. Crucially, the removal of information from the original model requires that the unlearned model makes identical predictions to a new model trained from scratch on the retained dataset (i.e., the dataset without the forget-set). Two important and different unlearning tasks studied in Machine unlearning literature are represented by class-removal and homogeneous sample removal scenarios.  In the former, the unlearned model has to discard entire classes.
A compelling use case for it can be found in the context of content moderation on social media platforms. These platforms often employ machine learning models to automatically detect and filter out content that violates their terms of service, such as hate speech, misinformation, or explicit material. As societal norms and regulatory requirements evolve, certain types of content that were previously deemed acceptable might need to be retrospectively removed from the platform and the training datasets of these models. Differently, in the homogeneous sampling removal tasks, the model unlearns a random subset of the training set. This case can be exemplified in the context of sensitive data management within healthcare predictive modeling. Healthcare AI systems are designed to predict patient outcomes based on a comprehensive set of electronic health records (EHR). Over time, certain patients might exercise their right to have their data removed from the dataset, either for privacy concerns or due to the deletion of incorrect or outdated information. 

Motivated by the growing need for more precise privacy-preserving unlearning methods, this paper introduces DUCK, a novel methodology that leverages metric learning. During the unlearning steps, DUCK initially computes centroids in the embedding space for each class in the dataset. Then, it minimizes the distance between the embeddings of forget-samples and the incorrect-class closest centroid. Importantly, DUCK can adapt to different unlearning scenarios such as class-removal or homogeneous sampling removal tasks. Overall, the main contributions of this paper can be summarized as:
\begin{itemize}
    \item Introduction of a novel machine unlearning method, Distance-based Unlearning via Centroid Kinematics \textbf{DUCK}, that employs metric learning to guide the removal of sample information from the model's knowledge. This involves directing samples towards the nearest incorrect centroid in the multidimensional feature space.
    
    \item Development of a novel metric, the Adaptive Unlearning Score (AUS), designed to quantify the trade-off between the forget-set accuracy and the overall test accuracy of the unlearned model.
    
    \item Extensive analysis of the unlearning mechanism employed in DUCK. Specifically, we investigated the modifications induced by DUCK into the feature space structure by analyzing the variation of the density of points (num. of points per unit of hyper-volume) for each class. Furthermore, we studied the unlearning mechanism of DUCK and the information removal procedure using Shapley values \cite{SHAP}, an explainable AI technique.
    
    \item Comprehensive experimental evaluations have been conducted on three publicly available datasets and against several related methods from the state-of-the-art (SOTA).
\end{itemize}
\section{Related Works}
\label{sec:related_works}

Machine unlearning is an emerging research area of artificial intelligence, enabling the removal of sensitive information from the weights of a model. This broad category of algorithms can be split into three main branches, as outlined in \cite{nguyen2022survey}: model-intrinsic approaches, data-driven approaches, and model-agnostic methods.

\noindent \textbf{(i) Model-Intrinsic Approaches.} This category encompasses algorithms that rely on specific architectural solutions to facilitate unlearning. These methods utilize sophisticated mechanisms designed to operate within the constraints of the chosen architecture. Depending on the task, various architectures may be employed to facilitate the unlearning process. For instance, in \cite{golatkar2020forgetting, golatkar2020eternal}, 2 unlearning methods are proposed for CNN architectures. In the first approach, the authors employed a technique based on the computation of the Fisher information matrix to eradicate knowledge of image instances from the model weights. Similarly, in the second approach, the authors introduced a CNN-based unlearning algorithm inspired by the neural tangent kernel. This algorithm models the dynamics of the weights and can eliminate information about forget data. Similarly, methodologies outlined in \cite{lin2023erm, pmlr-v119-guo20c}, employ specific strategies like weight masking and perturbation within CNNs for unlearning specific data samples. 

Furthermore, the interaction between architecture and the task has led the unlearning community to devise forgetting mechanisms specially tailored to the application's needs. Examples of this adaptation are visible in fields such as Multimodal Learning \cite{poppi2024removing, cheng2023multimodal}, Adversarial Learning \cite{tiwary2023adapt, Liu_2023_ICCV}, Graph Neural Networks \cite{cheng2023gnndelete,cong2023efficiently, pan2023unlearning}, Bayesian Models \cite{nguyen2020variational, fu2022knowledge, pmlr-v108-pearce20a}, and Statistical Learning algorithms \cite{liu2021revfrf,pmlr-v139-brophy21a,tarun2023deep}.

\noindent \textbf{(ii) Data-Driven Approaches.} 
These approaches are characterized by removing information about the forget samples from the model weights using the training, retain, and forget-sets. For instance, in \cite{setlur2022adversarial, sommer2020towards}, data transformations have been investigated as a useful approach to erase stored knowledge. Furthermore, within this broad category, gradient-based approaches represent a large subset. These methods assess the influence of the forget samples on the gradient updates and mitigate their impact using strategies like gradient inversion \cite{graves2021amnesiac,wu2020deltagrad}, gradient perturbation \cite{neel2021descent}, reservoir sampling \cite{neel2021descent}, and gradient re-weighting \cite{wu2022puma}. 

This category encompasses also methods tailored for Few-Shot unlearning \cite{yoon2022few, ramkumar2023learn} and Zero-Shot unlearning \cite{chundawat2023zero}. Additionally, methods based on random labels assignment \cite{hayase2020selective} and negative gradient updates\cite{golatkar2020eternal} fall in this category. The former, called Random Label, assigns to each forget sample a random label during the computation of the loss. The latter instead optimizes the weights of the network using the forget samples following the ascending direction of the gradient. A different branch of data-driven approaches is composed of the data-distributed based algorithms, such as SISA\cite{bourtoule2021machine}, where data and models are split among different nodes, and during the unlearning only the model belonging to the node that contains the forget data is manipulated.

\noindent \textbf{(iii) Model-Agnostic Approaches.} This group of algorithms comprises unlearning methods that can be applied independently to different Deep Neural Network architectures. For example, authors in \cite{felps2020class, Chen_2023_CVPR} introduced 2 techniques that shift the decision boundary by assigning incorrect classes to each forget-set sample. In the first study, this process is repeated until a Membership Inference Attack is no longer successful. In the latter, the model is finetuned on both the retain-set and the forget-set with the wrong labels.

In the studies of \cite{kurmanji2023towards,chundawat2023_badTeach}, the authors applied knowledge distillation to remove the knowledge associated with the forget-set. In \cite{kurmanji2023towards} authors maximize the distillation loss between the unlearning model and the original model whereas authors in \cite{chundawat2023_badTeach} distill the knowledge of a randomly initialized model into the unlearning model. A comparable approach was suggested in \cite{kim2022efficient}, where distillation was paired with contrastive labeling to accomplish efficient unlearning.

Other strategies include unlearning through model pruning \cite{jia2023model} and error-maximizing noise addition \cite{tarun2023fast}. In the former, it is demonstrated that model pruning can aid in the unlearning process, whereas in the latter, noise matrices are linked with forget classes corrupting those weights in the model that allows the recognition of the forget-set data.

\vspace{2mm}
In this paper, we introduce DUCK, a fast, efficient, and model-agnostic algorithm for approximate unlearning. DUCK partially aligns with Boundary Unlearning \cite{Chen_2023_CVPR} which modifies the forget class boundary manipulating the forget samples labels to induce unlearning. Conversely, our method operates directly within the feature space, driving the feature vectors of forget samples toward the nearest incorrect centroid. This strategy allows the model to remove the knowledge about the forget-set effectively, as demonstrated in Section \ref{sec:experiments} while preserving the knowledge about the retain-set. Furthermore, this approach is particularly crucial in scenarios where the forget-set consists of instances from multiple classes, which we refer to as the homogeneous removal scenario. In this setting, the unlearning of forget samples has to be targeted on sample-specific knowledge and preserve the original model generalization capabilities on never-seen data.

\section{Methods}
\label{sec:methods}
 In this section, we present a definition of the machine unlearning problem and how our solution, DUCK, can be applied to a general deep learning architecture to induce the forgetting of a subset of data from the training set.

\subsection{Preliminaries} 
In a general classification task, a dataset is defined by a set of tuples $\{x_i, y_i\}^{N}_{i=1}$ where $x_i$ denotes images, and $y_i$ represents their corresponding labels. The label values $y_i$ fall within the range $\{0, \dots, K-1\}$, where $K$ is the total number of classes in the dataset. Typically, this dataset is partitioned into two subsets: the training set $\mathcal{D}$ and the test set $\mathcal{D}^t$.
If there is a request for the removal of a particular class $C$ from $\mathcal{D}$, a machine unlearning algorithm must be applied to the original classification model $\Phi_\theta$ trained on $\mathcal{D}$. This approach aims to eliminate the information about class $C$ from the model weights $\theta$. We term this unlearning task as the Class-Removal (CR) scenario. In CR, the two sets $\mathcal{D}$ and $\mathcal{D}^t$ are divided into retain-sets ($\mathcal{D}_r$, $\mathcal{D}^t_r$) and forget-sets ($\mathcal{D}_f$, $\mathcal{D}^t_f$). The retain-sets contain all instances of images that should be preserved, denoted as $\mathcal{D}r=\{x^r_i, y^r_i\}_{i=1}^{N^{r}}$, with $y^r_i \in \{0, \dots, K-1\} \setminus \{C\}$. Conversely, the forget-sets contain only the images associated with the class label to be removed, i.e., $\mathcal{D}_f=\{x^f_i, y^f_i\}^{N^{f}}_{i=1}$ with $y^f_i \in \{C\}$. Ideally, through the utilization of these sets, the unlearned model $\Phi_{\theta}^{U}$ should achieve performance on the test set akin to that of an ``oracle'' model trained solely on the retain-set $\mathcal{D}_r$. This entails maximizing accuracy on $\mathcal{D}^t_r$ while being completely unaware of $\mathcal{D}^t_f$.

We also considered an alternative unlearning scenario defined as Homogeneous Removal (HR). This approach involves uniformly sampling from the dataset $\mathcal{D}$ to create a subset of data called $\mathcal{D}_f$, which comprises the samples intended for removal. Notably, in the HR scenario, the test set $\mathcal{D}^{t}$ is not divided into retain and forget subsets, as this would be superfluous. In fact, the unlearned model should preserve its generalization capabilities across classes that are modified by the removal of $\mathcal{D}_f$ samples. The remaining data from the training set are defined as $\mathcal{D}_r = \mathcal{D} \setminus \mathcal{D}_f$. Unlike CR, the HR scenario requires maintaining general knowledge about classes, resulting in a more targeted removal of information. Indeed, the primary goal of the unlearning algorithm is to erase specific samples $x_i \in \mathcal{D}_f$ from the training set, making them indistinguishable from images in $\mathcal{D}^{t}$, essentially treating $\mathcal{D}_f$ samples as never-seen-data. This degree of indistinguishability should also manifest in the accuracy on $\mathcal{D}_f$, which ideally should mirror the accuracy on $\mathcal{D}^t$. Moreover, it is fundamental that the original model's performance on the test set $\mathcal{D}^{t}$ does not deteriorate during unlearning.

\subsection{Proposed Method}
Given an original model $\Phi_{\theta}$, our approach aims to erase the information about the forget-set samples from the model weights $\theta$, directly influencing the feature vectors derived from $\Phi_{\theta}$. The network $\Phi_{\theta}$ can be represented as $ \Gamma_{\theta} \circ \Psi_{\theta} $, where $\Psi_{\theta}$ constitutes the backbone of the network, typically encompassing convolutional layers in the case of a CNN, while $\Gamma_{\theta}$ represents the last fully-connected layer of the network (Fig. \ref{fig:scheme}A). The unlearning process in DUCK consists of a two-phase optimization procedure each characterized by either a high or low level of forgetting. In the high-forget regime, DUCK eliminates knowledge about the forget set from the DNN. Conversely, during the low-forget regime, DUCK restores the retain performance without reacquiring knowledge about the forget set.

Before these two phases, the algorithm calculates the centroid vector $c_i$ in the embedding space for each class in $\mathcal{D}_r$. This involves computing $c_i$ as the mean embedding vector, as detailed in eq. \ref{eq:centroids}.
\begin{equation}
c_i = \frac{1}{N_i} \sum_{j=0}^{N_i-1} \Psi_{\theta}(x_j)
\label{eq:centroids}
\end{equation}
where $N_i$ is the number of samples in $\mathcal{D}_r$ for the $i^{\text{th}}$-class.

\minisection{Optimization Procedure.}After retrieving the centroids, the optimization procedure involves calculating the forget-loss from forget-set batches and the retain-loss from retain-set batches.
For each sample in the forget-set batch $(x_j, y_j)$, DUCK selects the centroid in $\{c_i\}_{i=0}^{K-1}$ belonging to a class different from $y_j$ that minimizes the pairwise cosine distance $d$ from $\Psi_{\theta}(x_j)$, as defined in eq. \ref{eq:distance} (Fig. \ref{fig:scheme}A-B).
\begin{equation}
    c^*_j = \argmin_{c_i} d(\Psi_{\theta}(x_j),c_i)
    \label{eq:distance}
\end{equation}
This stage of the algorithm is crucial as it determines the optimal directions for moving the embeddings of forget-samples. Subsequently (Fig. \ref{fig:scheme}C), the loss is computed by assessing the cosine distances between the embeddings $\Psi_{\theta}(x_j)$ and the selected closest centroid $c^*_j$ as follows:
\begin{equation}
    \mathcal{L}_{FGT} = \frac{1}{N_{\text{batch}}} \sum_{j=0}^{N_{\text{batch}}-1} 1-\frac{\Psi_{\theta}(x_j) \cdot c^*_j}{||\Psi_{\theta}(x_j)|| \;||c^*_j||}
\label{eq:loss_fgt},
\end{equation} 
where $N_{\text{batch}}$ is the number of forget samples in the batch. Minimizing this loss involves moving the extracted embeddings of forget samples away from the correct class centroid and toward the closest incorrect class centroid (the impact of the unlearning algorithm on the position of the embeddings in the latent space is discussed in Sec \ref{subsec:manifold}). 

The retain-loss considers the entire network $\Phi_{\theta} = \Gamma_{\theta} \circ \Psi_{\theta}$ and entails computing the cross-entropy loss of the retain-set batch samples $(x_j, y_j)$, as expressed in eq. \ref{eq:loss_ret}.
\begin{equation}
    \mathcal{L}_{RET} = -\frac{1}{N_{\text{batch}}}\sum_{j=0}^{N_{\text{batch}}-1} \sum_{i = 0}^{K-1} y_i log(p_{i,j})
    \label{eq:loss_ret}
\end{equation}
where $K$ is the number of classes, $N_{\text{batch}}$ is the number of retain samples in the batch, $y_i$ the true label, and $p_{i,j}$ is the probability of the $i^{th}$ class obtained from $\texttt{softmax}\left[\Gamma_{\theta} \circ \Psi_{\theta} (x_j)/T\right]$ (T is the temperature that allows calibrating the logits). This loss serves as a knowledge-preserving method for the unlearning algorithm since it counterbalances the effect of the $\mathcal{L}_{FGT}$ on the backbone $\Psi_{\theta}$ weights (in Sec \ref{subsec:ablation} we analyze the impact of the components of the loss function in an ablation study).

Finally, the overall loss is computed as a weighted combination of $\mathcal{L}_{FGT}$ and $\mathcal{L}_{RET}$ eq. \ref{eq:loss}
\begin{equation}
    \mathcal{L} = \lambda_1 \mathcal{L}_{FGT} + \lambda_2 \mathcal{L}_{RET}
    \label{eq:loss}
\end{equation}
where $\lambda_1$ and $\lambda_2$ are hyperparameters. 

\minisection{Forget Regimes.} During the high-forget regime phase, the number of epochs in the optimization procedure is dynamic and based on the accuracy $\mathcal{A}_f$ on the train forget-set $\mathcal{D}_f$ computed after each epoch.  The unlearning process is halted as soon as the computed $\mathcal{A}_f$ drops below a predefined optimal value $\mathcal{A}^{\star}_f$. If this condition is not met, the unlearning automatically concludes after 10 epochs in every experiment. In CR, the optimal accuracy threshold $\mathcal{A}^{\star}_f$ is adjusted to 0\% plus a tolerance margin ($\delta$), acknowledging the practical difficulty of achieving an exact 0\% accuracy. This tolerance $\delta$ is set to 1\% in CR experiments to account for this challenge. In contrast, in the HR scenario, the optimal value for $\mathcal{A}_f$ aligns with the original model test accuracy $\mathcal{A}^t_{Or}$. The objective in this case is to forget $\mathcal{D}_f$ and at the same time maintain the generalization capability of the model. If  $\mathcal{D}_f$ are properly forgotten these data can be interpreted as never-seen-data and this reflects in terms of accuracy in having $\mathcal{A}_f \approx \mathcal{A}^t$ where the latter term should be close as possible to $\mathcal{A}^t_{Or}$. In the low-forget regime phase, the forget loss hyperparameter $\lambda_2$ is scaled by 0.1 in CR and 0.3 in HR, to stabilize the forget-set feature vectors while DUCK restores any performance loss induced by the information removal procedure. During this phase, the number of epochs is fixed to 2 in every experiment.

\section{Experimental Results}
\label{sec:experiments}
In this section, we present the experimental results obtained on three distinct datasets sourced from SOTA research. We compared our method with others from the SOTA, focusing on CR and HR scenarios. Subsequently, we provide various analyses conducted to evaluate the effectiveness of our unlearning approach.

\subsection{Experimental Setting}
\label{subsec:exp_set}
For our experiments we analyzed the CIFAR10\cite{krizhevsky2009learning}, CIFAR100\cite{krizhevsky2009learning}, and TinyImagenet\cite{Le2015TinyIV} datasets. Details about the datasets are provided in the Supp. Materials. We compared DUCK with the following methods and baselines on the CR and HR scenarios:\\
\textbf{Original}: Original model trained on $\mathcal{D}$ without any unlearning procedure. This is the original model from which every unlearning method begins.\\
\textbf{Retrained}: This is the ``oracle'' model, trained for 200 epochs on $\mathcal{D}_r$, without any knowledge of the forget-set $\mathcal{D}_f$.\\
\textbf{Finetuning\cite{golatkar2020eternal}}: The original model is fine-tuned on the retain-set $\mathcal{D}_r$ for 30 epochs, employing a high learning rate to remove knowledge from the forget-set and optimize accuracy on the retain-set. It's crucial to note that altering the epoch count significantly impacts fine-tuning performance. A small number of epochs results in inefficient unlearning, while prolonged fine-tuning essentially retrains the model. Therefore, the original model undergoes fine-tuning for an adequate number of epochs ($\ll$ epochs of retrain) to ensure the erasure of $\mathcal{D}_f$, as delineated in prior studies \cite{Chen_2023_CVPR,lin2023erm}.\\
\textbf{Negative Gradient (NG)\cite{golatkar2020eternal}}: the original model is fine tuned on the forget-set $\mathcal{D}_f$ following the negative direction of the gradient descent. The unlearning is stopped when the forget accuracy on the forget test set is 0.\\
\textbf{Random Label (RL)\cite{hayase2020selective}}: the original model is fine-tuned with the forget-set $\mathcal{D}_f$ where labels are changed with a random label using cross-entropy loss function. The unlearning is stopped when the forget accuracy on the forget test set is 0.\\
\textbf{Boundary Shrink (BS) and Expanding (BE)\cite{chen2023boundary}}: in BS, first the closest and incorrect class label is found and assigned to each forget sample, then the model is fine-tuned with both the retain and the forget (with the changed labels set). In BE instead, an extra shadow class is assigned to the forget sample and then the model is fine-tuned as in BS. \\
\textbf{ERM-KTP (ERM)\cite{lin2023erm}}: this method alternates the Entanglement-Reduced Mask and Knowledge Transfer and Prohibition phases to remove the information about the forget-set and maximize the accuracy on the retain-set.\\
\textbf{SCRUB\cite{kurmanji2023towards}}: This approach utilizes a distillation mechanism, called ``max-steps'', to discard the information from the forget-set, by pushing the predictions of the student network (unlearned model) away from those of the teacher network (original model). However, SCRUB switches between max-steps and a specific recovery process for the retain-set called ``min-steps'' based on the combination of distillation and cross-entropy losses applied on the retain set. Furthermore, the authors performed a limited number of ``min-steps'' to recover the lost knowledge associated with the retained data.\\
\textbf{L1-Sparse\cite{jia2024_l1sparse}}:  In this work, the authors explored model pruning before applying the unlearning strategy. They found out that a sparse model is more prone to better unlearning than a dense model. They applied this to a variety of unlearning methods showing the results in the pruned and dense cases. Among the different algorithms involved, we selected the L1-Sparse Method.\\
\textbf{SSD\cite{foster2023SSD}}: Selective Synaptic Dampening (SSD) utilizes the Fisher information matrix derived from both the training and forget data to identify which DNN parameters are crucial to the forget-set. Subsequently, it accomplishes forgetting by attenuating these identified parameters in proportion to their significance to the forget-set compared to the entire training data.\\
\textbf{Bad-Teacher\cite{chundawat2023_badTeach}}: in this approach, like SCRUB, distillation serves as the primary method for erasing and retaining information from the forget- and retain- sets, respectively. Either a randomly initialized model (referred to as the ``bad-teacher model'') is distilled into the model to be unlearned (the student model) using the forget-set, or the original model is distilled into the student using the retain-set.

During our experiments with DUCK, we set the batch size to 1024 for CIFAR10, CIFAR100, and TinyImageNet in both CR and HR scenarios. 
For all the experiments we adopted Adam optimizer \cite{loshchilov2017decoupled}, with weight decay set to $5\times10^{-4}$. We reported in Supp. Materials the values used for $\lambda_1$, $\lambda_2$, and the learning rate in each experiment. Moreover, we also provide the hyperparameters used in experiments for the other SOTA methods. For DUCK and all the considered methods, we use \texttt{Resnet18} as architecture. 
We compared the performance of the considered methods in terms of accuracies on the forget and retain-set both for test data. In the context of the CR scenario, we denoted $A^t_r$ and $A^t_f$  as the accuracies on the retain and forget-set, respectively, derived from the test set. In the HR scenario, the data from the test set are not split, hence we defined $A_r$, $A_f$, and $A^t$ as the accuracies on the retain, forget, and test set. Moreover, we defined a new metric called \textbf{Adaptive Unlearning Score} \textbf{(AUS)} to better capture the trade-off between the forget-set accuracy and the overall test accuracy of the unlearned model.
\begin{figure}
    \centering
    \includegraphics{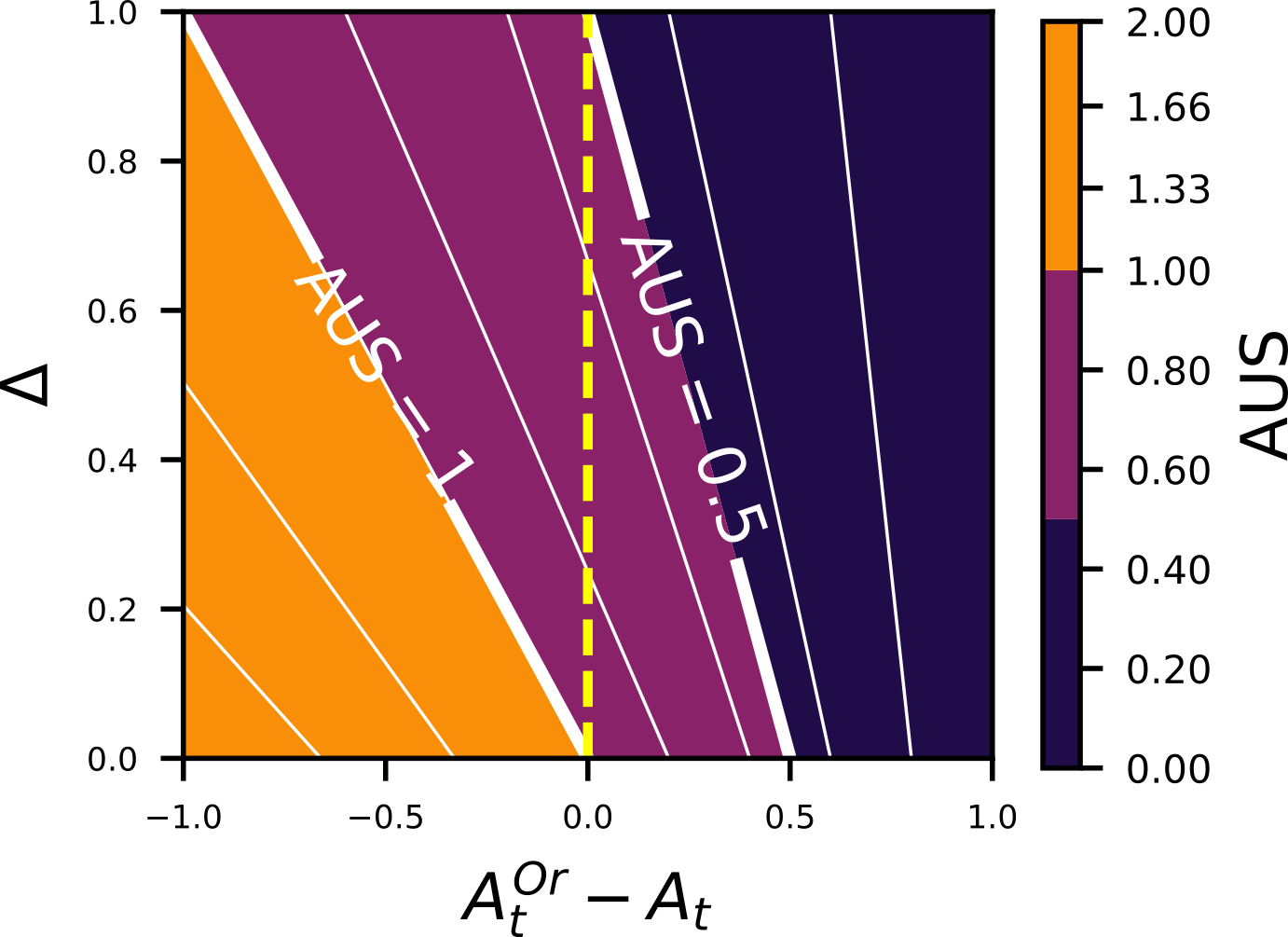}
    \caption{Contour plot illustrating the Adaptive Unlearning Score (AUS) as a function of $\Delta$ and the difference between the original and unlearned model's test accuracies ($A^{Or}_t - A_t$). }
    \label{fig:AUS}
\end{figure}

\minisection{Adaptive Unlearning Score.} This metric addresses the need for a measure that takes into account the performance of an unlearned model on both the retain-set and the forget-set. The metric is defined in eq. \ref{eq:AUS}:
\begin{equation}%
    \text{AUS} = \frac{1-(\mathcal{A}^{Or}_{t}-\mathcal{A}_{t})}{1 + \Delta}, \hspace{1em}    \Delta = 
    \begin{cases}
      | 0 - \mathcal{A}_{f}| & \text{if CR}\\
      | \mathcal{A}_{t} - \mathcal{A}_{f}| & \text{if HR} 
    \end{cases},
    \label{eq:AUS}
\end{equation}
where $A_{t}$ and $A_{f}$ are respectively the accuracy on the test set and forget-set of the unlearned model, and $A^{Or}_t$ is the accuracy of the original model on the test samples (to prevent excessive notation in the formula, for the CR scenario $\mathcal{A}_f = \mathcal{A}^t_{f}$, $\mathcal{A}^{Or}_t = \mathcal{A}^{t,Or}_r$ and $\mathcal{A}_t=\mathcal{A}^t_r$ ). The AUS reflects the balance between two compelling needs: maintaining high test accuracy while addressing the unlearning task. In Eq. \ref{eq:AUS}, the numerator measures how close the test accuracy $\mathcal{A}^t$ of the unlearned model is to the accuracy of the original model. Simultaneously, the denominator measures how close the forget accuracy is to the scenario-dependent target accuracy. AUS provides a significant advantage by facilitating the comparison between cases that are typically challenging to assess directly, such as when both the test accuracy ($\mathcal{A}^t$) and forget-set accuracy ($\mathcal{A}_f$) are either high or low. By incorporating both $\mathcal{A}_f$ and $\mathcal{A}_t$ into a single, comprehensive score, AUS enables a straightforward ranking of different unlearning methods.

The contour plot depicted in Figure \ref{fig:AUS} provides a detailed visualization of the AUS as it varies with $\Delta$ and the difference between the test accuracy of the original model and that of the unlearned model ($A^{Or}_t - A_t$). 
The yellow dashed line in the figure represents the scenario where $A^{Or}_t = A_t$, meaning the test accuracy of the unlearned model equals that of the original model. This line marks the theoretically accessible values of AUS for the original model (AUS$^{Or}$), before any unlearning occurs. AUS$^{Or}$ is typically low and very close to the top of the plot in CR ($\mathcal{A}_{f}^{Or}\simeq\mathcal{A}_r^{Or}\rightarrow \Delta\gg0$). Conversely, AUS$^{Or}$ is higher in HR, where $\Delta$ is proportional to the degree of model overfitting and is expected to be not excessively large. 

The dark blue region, indicating an AUS below 0.5, highlights scenarios where the unlearning process has adversely affected the model's performance. This reduction in AUS could stem from a decrease in $\mathcal{A}_t$ accompanied by an insufficient decline in $\mathcal{A}_f$, suggesting that the unlearning process is not optimally targeting the forgettable information. Conversely, the orange area, representing an AUS greater than 1, indicates instances where the unlearned model significantly outperformed the original model on the retain-set. Such an increase in AUS post-unlearning could be due to two main reasons: \textbf{\emph{i})} The result might represent a statistical fluctuation; hence, repeating the experiment could yield values closer to an AUS of 1. \textbf{\emph{ii})} It is possible that the original model was not optimally trained, suggesting that during the unlearning process, the model may have exploited previously overlooked or underutilized data, thus improving its performance.

\begin{table*}[!ht]
\caption{Comparison between DUCK and SOTA methods for the CR scenario. The metrics reported here $\mathcal{A}^t_r$, $\mathcal{A}^t_f$, and AUS are mean $\pm$ std over ten runs, each with a different class has been removed. $^{\dag}$ Standard deviations are not available for ERM  since the code released in \cite{lin2023erm} does not allow to remove classes arbitrarily; the results are reported for class 0.}
\label{tab:oneclass}
\resizebox{1.\linewidth}{!}{\begin{tabular}{l|ccc|ccc|ccc}
\toprule
                  &  \multicolumn{3}{c|}{CIFAR10}  & \multicolumn{3}{c|}{CIFAR100}  & \multicolumn{3}{c}{TinyImagenet}  \\
                  & $\mathcal{A}_r^t$ & $\mathcal{A}_f^t$& AUS & $\mathcal{A}_r^t$ & $\mathcal{A}_f^t$& AUS & $\mathcal{A}_r^t$ & $\mathcal{A}_f^t$& AUS \\
\midrule
Original          & 88.64 (00.63) & 88.34 (05.62) & 0.531 (0.005) & 77.55 (00.11) & 77.50 (02.80) & 0.563 (0.009) & 68.26 (00.08) & 64.60 (15.64) & 0.607 (0.058)\\
Retrained         & 88.05 (01.28) & 00.00 (00.00) & 0.994 (0.014) & 77.97 (00.42) & 00.00 (00.00) & 1.004 (0.022) & 67.67 (01.00) & 00.00 (00.00) & 0.993 (0.010)\\
Fine Tuning       & 87.93 (01.14) & 00.00 (00.00) & 0.993 (0.013) & 77.31 (02.17) & 00.00 (00.00) & 0.998 (0.022) & 67.89 (00.24) & 00.00 (00.00) & 0.994 (0.010)\\
\hline
Neg. Grad.        & 76.27 (03.27) & 00.56 (00.12) & 0.871 (0.033) & 62.84 (06.13) & 00.50 (00.50) & 0.849 (0.061) & 60.09 (02.58) & 00.60 (01.35) & 0.911 (0.028)\\
Rand. Lab.        & 65.46 (07.59) & 00.83 (00.44) & 0.762 (0.076) & 55.31 (07.06) & 00.40 (00.70) & 0.774 (0.070) & 43.29 (10.10) & 01.20 (01.03) & 0.740 (0.100)\\

Boundary S.       & 83.81 (02.29) & 13.64 (03.57) & 0.840 (0.543) & 55.07 (11.85) & 03.54 (02.26) & 0.749 (0.116) & 55.98 (03.33) & 04.25 (02.12) & 0.840 (0.036)\\
Boundary E.       & 82.36 (02.39) & 13.34 (03.21) & 0.827 (0.032) & 55.18 (11.98) & 03.50 (02.32) & 0.750 (0.117) & 55.92 (03.45) & 02.60 (02.50) & 0.853 (0.040)\\
SCRUB             & 87.88 (01.18) & 00.00 (00.00) & 0.992 (0.013) & 77.29 (00.25) & 02.00 (05.35) & 0.977 (0.051) & 68.15 (00.25) & 01.20 (03.79) & 0.986 (0.037)\\
L1-Sparse         & 85.21 (09.41) & 04.88 (02.18) & 0.862 (0.023) & 62.63 (07.56) & 01.22 (00.53) & 0.879 (0.018) & 63.76 (01.29) & 04.81 (01.54) & 0.901 (0.011)\\
ERM-KTP$^\dagger$ & 80.31 (-)     & 00.00 (-)     & 0.917 (-)     & 58.96 (-)     & 00.00 (-)     & 0.814 (-)     & 49.45 (-)     & 00.00 (-)     & 0.811 (-)    \\
Bad Teacher       & 88.89 (00.87) & 02.37 (05.86) & 0.982 (0.057) & 77.19 (00.19) & 06.00 (10.09) & 0.940 (0.089) & 64.09 (00.78) & 11.20 (09.04) & 0.862 (0.070)\\
SSD               & -             & -             & -             & 74.02 (03.46) & 00.00 (00.00) & 0.932 (0.012) & -             & -             & -            \\
DUCK              & 88.46 (00.95) & 00.00 (00.00) & \textbf{0.998 (0.011)} & 77.18 (00.28) & 00.00 (00.00) & \textbf{0.996 (0.003)} & 68.47 (00.33) & 00.00 (00.00) & \textbf{1.002 (0.003)}\\

\bottomrule
\end{tabular}}
\end{table*}
\begin{table*}[!ht]
\caption{Comparison between DUCK and SOTA methods for the HR scenario. The metrics $\mathcal{A}_f$, $\mathcal{A}^t$, and AUS reported here are mean $\pm$ std over ten runs, where for each run a different seed has been used and a different 10\% of the data is forgotten.}
\label{tab:homogeneous}
\resizebox{1.\linewidth}{!}{\begin{tabular}{l|ccc|ccc|ccc}
\toprule
                        & \multicolumn{3}{c|}{CIFAR10}                                                 & \multicolumn{3}{c|}{CIFAR100}                                                & \multicolumn{3}{c}{TinyImagenet}                                                    \\
                  & $\mathcal{A}^t$ & $\mathcal{A}_f$ &  AUS & $\mathcal{A}^t$ &$\mathcal{A}_f$ & AUS & $\mathcal{A}^t$ & $\mathcal{A}_f$ &  AUS\\
\midrule
Original        & 88.54 (00.25) & 99.49 (00.08) & 0.900 (0.004) & 77.23 (00.53) &99.63 (00.10)  & 0.814 (0.006)  &68.08 (00.39) & 84.83 (00.44) & 0.854 (0.007) \\

Retrained       & 84.13 (00.59) &84.56 (00.72)  & 0.950 (0.011) & 77.27 (01.03) &76.87 (00.95)  & 0.993 (0.017)  &63.45 (02.34) & 63.24 (02.58) & 0.950 (0.040)\\

Fine Tuning     & 85.70 (00.48) &88.66 (00.44)  & 0.942 (0.008) & 72.06 (00.52) &74.97 (00.72)  & 0.918 (0.009)  & 67.03 (00.45) &70.52 (00.65) & 0.937 (0.011)\\
\hline
Neg. Grad.      & 79.35 (00.85) & 87.11 (00.86) & 0.841 (0.012) & 60.83 (00.77) & 76.77 (00.57) &  0.718 (0.009) & 54.83 (00.81) & 67.27 (00.59) & 0.770 (0.011) \\
Rand. Lab.      & 77.28 (01.37) & 87.07 (01.20) & 0.807 (0.018) & 60.51 (00.82) & 77.18 (00.50) & 0.711 (0.009)  & 54.34 (00.85) & 67.65 (00.45) & 0.760 (0.011) \\
SCRUB           & 78.02 (07.80) &82.53 (10.80)  & 0.854 (0.132) & 73.33 (00.85) &92.46 (01.50)  & 0.804 (0.014)  & 68.52 (00.40) &83.14 (00.53) & 0.875 (0.008)\\

L1-Sparse       & 93.05 (00.53) &99.06 (00.65)  & 0.927 (0.011) & 52.26 (01.65) &66.89 (02.34)  & 0.690 (0.024)  & 63.76 (01.19) &84.30 (01.04) &0.794 (0.032)\\
Bad Teacher     & 86.66 (00.44) & 92.91 (00.57) & 0.922 (0.008) & 73.57 (00.61) & 86.57 (00.57)  & 0.851 (0.006) & 67.41 (00.49) & 80.28 (00.81)  & 0.876 (0.006)\\
SSD             & - &- &  -                                     & 75.41 (02.26)  & 97.70 (02.11) &  0.805 (0.013)  &  - & -   & -\\
DUCK            & 87.81 (00.32) &87.28 (00.55)  & \textbf{0.986 (0.008)} & 76.66 (00.63) &75.72 (01.93)  & \textbf{0.982 (0.021)}  & 67.55 (00.41) & 67.24 (00.64)& \textbf{0.990 (0.010)} \\

\bottomrule
\end{tabular}}

\end{table*} 

\subsection{Class-Removal Scenario}
\label{sec:compar}
In the CR scenario, the main purpose of the unlearning algorithm is to remove specific class-related information from the parameters of the original model; therefore, the forget-set only includes elements from a single class. Table \ref{tab:oneclass} presents the results of DUCK and the competitors from the SOTA in the CR scenario for CIFAR10, CIFAR100, and TinyImagenet. The reported values represent the means and standard deviations from 10 different runs, with a different class making up the forget-set each time. Notably, our method stands out as the only approach capable of erasing class-specific information (i.e. $\mathcal{A}^t_f\approx 0$) while preserving the highest possible accuracy on the retain test set. Moreover, DUCK exhibits performance closely aligned with more time-consuming techniques such as FineTuning and Retrained (see Sec. \ref{subsec:time} for unlearning time analysis). Nevertheless, our approach is more efficient than these two methods, which are considered upper-bound models as they are optimized on the entire retain-set over numerous training iterations. These results are validated by the AUS scores, confirming the practical effectiveness of our approach.

\subsection{Homogenous-Removal Scenario}
\label{subsec:dataremoval}
While the most studied unlearning scenario in the literature is CR, we have also extensively tested DUCK on the HR scenario, which constitutes a fundamental open problem and a relevant use case for privacy-compliant applications. In HR problems, unlike CR, the information about all classes must be preserved as much as possible while simultaneously removing information about the forget-set. This configuration prohibits unlearning methods from exploiting an inductive bias regarding entire class removal, forcing them to operate at the level of a single forget sample. Indeed, removing the information about forget classes leads to complete failure in the HR scenario, as $\mathcal{D}_f$ and $\mathcal{D}^t$ share the same classes. The closest-centroid procedure in DUCK plays a crucial role in overcoming this difficulty. Minimizing the distance between $\Psi_{\theta}(x_j)$ and $c^*_j$ (Eq.\ref{eq:loss_fgt}) enables DUCK to operate on a single forget-sample $x_j$ rather than the entire class of $x_j$. Thanks to these characteristics, DUCK can be applied to unlearning problems regardless of the type of scenario. 

We present the results for this scenario in Table \ref{tab:homogeneous}, where for each of the 10 seeds, a different 10\% of the original dataset constitutes the forget-set. We have excluded from the comparison the methods that are by construction not suitable for this scenario. Overall, DUCK outperforms both the unlearning methods from SOTA and Finetuning and is comparable to the Retrained model over all the considered datasets. This result underscores the flexibility of DUCK in addressing HR scenarios and confirms the results obtained in CR scenarios. Moreover, not only does DUCK achieve results comparable to the model Retrained, but also it is on average $\approx20$ times faster and represents a suitable algorithm for unlearning problems.

\subsection{Membership Inference Attack}
\label{sec:mia}
Accuracy-based measures are crucial for monitoring the performance of the unlearned model. However, even if the values of these metrics are similar to those of the retrained model, they do not guarantee that information regarding the membership of forget data is erased (i.e., a forget-sample even if it cannot be classified correctly by the unlearned model, it could still be properly identified as a training set sample). Therefore, a Membership Inference Attack \cite{mia1,Carlini22,chen2023boundary} is fundamental to verify that forget set data cannot be identified as training set samples and consequently the effectiveness of the unlearning algorithm in scenarios such as CR and HR. For this reason, we applied the \emph{offline} Likelihood Ratio Attack (LIRA) algorithm proposed in \cite{Carlini22} in our unlearning scenarios. Given a target forget sample $(x,y)$, offline LIRA trains $N=128$ shadow models, on random samples (50\% of the total) from the original $\mathcal{D}$, making sure $(x,y)$ is left out.
Thus, the attack computes the confidence of each shadow model on $x$. In this case, the confidence is computed as the logit-scaled confidence (Eq. \ref{eq:confidence}).
\begin{equation}
    \begin{split}
        \alpha(x) &= log\left(\frac{ e^{-\mathcal{L}(\Phi_{\theta}(x),y)}}{1- e^{-\mathcal{L}(\Phi_{\theta}(x),y)}}\right)
    \end{split}
    \label{eq:confidence}
\end{equation}
where $\mathcal{L}$ is the cross-entropy loss. Then, the attacker parameterized the distribution of confidences as a Gaussian distribution $\mathcal{N}(\mu,\sigma)$ where the mean $\mu$ and variance $\sigma$ are estimated from the shadow models confidences samples. Finally, the attack measures the probability of observing confidence as high as the target model’s under the null hypothesis that the target point $(x, y)$ is a non-member of the target model training set eq. \ref{eq:mia_eq}.
\begin{equation}
\begin{split}
    &P(x\in \mathcal{D}_f) = 1-P(Z > \alpha(x))= \\
    =1-\int_{\alpha(x)}^{\infty} &\mathcal{N}(x;\mu,\sigma)\,dx =\frac{1}{2} \left( 1 + \text{erf}\left(\frac{\alpha(x)-\mu}{\sigma\sqrt{2}}\right)\right)
    \end{split}
    \label{eq:mia_eq}
\end{equation}
where ``erf'' is the Gauss error function and $Z\sim \mathcal{N}(\mu,\sigma)$. 
Overall, the larger the attacked model’s confidence is compared to $\mu$, the higher the likelihood that the query sample is a training set member.

We applied this MIA to evaluate the efficacy of DUCK in unlearning, comparing it against the original model and the retrained model. , using CIFAR10 as the dataset. The detailed outcomes are presented in Tab. \ref{tab:mia} (see Supp. Materials Figure 2 for receiver operating characteristic curve). Following the approach suggested by \cite{Carlini22}, we highlight the importance of measuring True Positive Rate (TPR) at low False Positive Rate (FPR) levels, since TPR metrics at FPRs exceeding 50\% lack practical relevance from an adversarial standpoint. In this analysis, we considered the TPR at different FPR ($1\%$ and $10^{-1}\%$) and the Area Under the Curve (AUC) score using as query samples both the forget-set $\mathcal{D}_f$ and either the forget test set $\mathcal{D}_f^t$ for CR or the test set $\mathcal{D}^t$ data for HR. Overall, offline LIRA results demonstrate how DUCK can remove the membership information about the forget set achieving  TPR at low FPR compatible with the retrained model.

\begin{table}
    \centering
    \caption{Results of the offline LIRA\cite{Carlini22} applied to CIFAR10 in CR and HR scenarios. For both scenarios, we reported the true positive rate (TPR) at 1 and 0.1 $\%$ of false positive rate (FPR) and AUC. Scores are reported as mean $\pm$ std across 5 forget classes$=0,1,2,3,4$ in CR and for 5 seeds in HR.}
    \resizebox{1.\linewidth}{!}{\begin{tabular}{lccc|ccc}
\toprule
              & \multicolumn{3}{c|}{CR}                       &\multicolumn{3}{c}{HR} \\
              & TPR $@ 1\%$  & TPR $@ 10^{-1}\%$ &AUC         & TPR $@ 1\%$ & TPR $@ 10^{-1}\%$ &AUC \\
\midrule
Or.     & 12.0 (03.80)       & 6.6 (03.1)  &0.655(0.056)     &7.1 (3.0)$\%$ & 02.2 (01.1)\%& 0.617(0.025) \\
Rt.     & 01.0 (00.8)       & 0.1 (0.08)  &0.500(0.004)     &1.0 (0.8)\%   & 0.1 (0.08)\%& 0.502 (0.004)\\
DUCK    & 01.0 (00.8)       & 0.1 (0.08)  &0.500(0.001)     &1.0 (0.8)\%   & 0.1 (0.08)\%& 0.506 (0.004) \\

\bottomrule
\end{tabular}}

    \label{tab:mia}
\end{table}

\begin{table}
    \centering
    \caption{Results of the ablation study on the CIFAR100 and TinyImagenet datasets in the CR and HR scenarios. The metrics reported here are mean over 10 runs, each run using a different forget class for CR or a different seed for HR.}
    \resizebox{1.\linewidth}{!}{\begin{tabular}{cccccccc}
\toprule
   \multicolumn{2}{l}{} & $\mathcal{L}_{\text{FGT}}$ & $\mathcal{L}_ {\text{RET}}$ & $\mathcal{A}^t_r$ & $\mathcal{A}^t_f$  & AUS   & Un. Time[s] \\
\midrule
\multirow{8}{*}{\rotatebox[]{90}{CR}}&\multirow{4}{*}{\rotatebox[]{90}{CIFAR100}}  &
     \xmark & \xmark & 77.55 (00.11) & 77.50 (02.80) & 0.563 (0.009) & -  \\
 & & \cmark & \xmark & 60.24 (05.51) & 00.90 (01.60) & 0.819 (0.056) & 5  \\
 & & \xmark & \cmark & 78.04 (00.12) & 39.00 (20.01) & 0.723 (0.104) & 142\\
 & & \cmark & \cmark & 77.18 (00.28) & 00.00 (00.00) & 0.996 (0.003) & 31 \\
\cmidrule{2-8}
&\multirow{4}{*}{\rotatebox[]{90}{TinyImg.}} & 
     \xmark & \xmark & 68.26 (00.08) & 64.60 (15.64) & 0.607 (0.058) &   -  \\
 & & \cmark & \xmark & 15.86 (05.82) & 02.40 (02.95) & 0.464 (0.058) &  12  \\
 & & \xmark & \cmark & 68.28 (00.13) & 06.60 (09.75) & 0.936 (0.080) &  980 \\
 & & \cmark & \cmark & 68.47 (00.33) & 00.00 (00.00) & 1.002 (0.003) &  186 \\
\midrule
\multicolumn{2}{l}{} & $\mathcal{L}_{\text{FGT}}$ & $\mathcal{L}_ {\text{RET}}$ & $\mathcal{A}^t$ & $\mathcal{A}_f$  & AUS & Un. Time[s] \\
 \midrule
\multirow{8}{*}{\rotatebox[]{90}{HR}}&\multirow{4}{*}{\rotatebox[]{90}{CIFAR100}}  & 
     \xmark & \xmark & 77.23 (00.53) & 99.63 (00.10) & 0.814 (0.006)  &   -  \\
 & & \cmark & \xmark & 08.77 (00.79) & 02.86 (00.37) & 0.295 (0.008) & 14\\
 & & \xmark & \cmark & 77.67 (00.53) & 98.93 (00.17) & 0.826 (0.006) & 318 \\
 & & \cmark & \cmark & 76.66 (00.63) & 75.72 (01.93) & 0.982 (0.021)  & 95  \\

\cmidrule{2-8}
&\multirow{4}{*}{\rotatebox[]{90}{TinyImg.}}       &
     \xmark & \xmark & 68.08 (00.39) & 84.83 (00.44) & 0.854 (0.007) &   -      \\
 & & \cmark & \xmark & 00.75 (00.27) & 00.75 (00.24) & 0.325 (0.006) & 89 \\
 & & \xmark & \cmark & 67.21 (00.57) & 79.67 (00.61) & 0.880 (0.009) & 2330 \\
 & & \cmark & \cmark & 67.55 (00.41) & 67.24 (00.64) & 0.990 (0.010) & 380  \\
\bottomrule
\end{tabular}}

    \label{tab:ablation}
\end{table}
\subsection{Ablation Study}
\label{subsec:ablation}
To investigate the contribution of the different components of DUCK we performed an ablation study, on CIFAR100 and TinyImagenet, selectively deactivating $\mathcal{L}_{RET}$ or $\mathcal{L}_{FGT}$ in the CR and HR scenarios (Table \ref{tab:ablation}) while keeping the remaining hyperparameters fixed. The first row of the Table reports the results of the original model without any unlearning procedure. When removing $\mathcal{L}_{RET}$ we observed a great reduction in $\mathcal{A}^t_r$ for CR ($\mathcal{A}^t$ for HR) in both CIFAR100 and TinyImagenet. This effect is even more pronounced in HR since in this scenario the closest-centroid mechanism affects the knowledge about each class. This result underlines the importance of the retain-set loss which counter-balance the effect of our closest-centroid mechanism to remove the information about forget-samples. Removing $\mathcal{L}_{FGT}$ essentially translates to tuning the original model on the retain-set using the stopping mechanism described in Sec \ref{sec:methods}. Nevertheless, the stopping condition was never reached for either CR or HR, and the original model was therefore tuned for 10 (high-forget regime) plus 2 (low forget-regime) epochs. This ablation prevents the unlearning algorithm from efficiently erasing the knowledge of the forget sample from the original model as highlighted by the high $\mathcal{A}_f^t$ in CR or $\mathcal{A}_f$ in HR. This effect results in a low AUS score even if the $\mathcal{A}^t_r$ or $\mathcal{A}^t$ are compatible with the original model test accuracy.

Overall these results confirm the importance of using both the components of the loss function for maintaining high performance while effectively removing data belonging to the forget-set.

\subsection{Multiple Class-Removal Scenario}
\begin{figure}
    \centering
    \includegraphics[width=.9\linewidth]{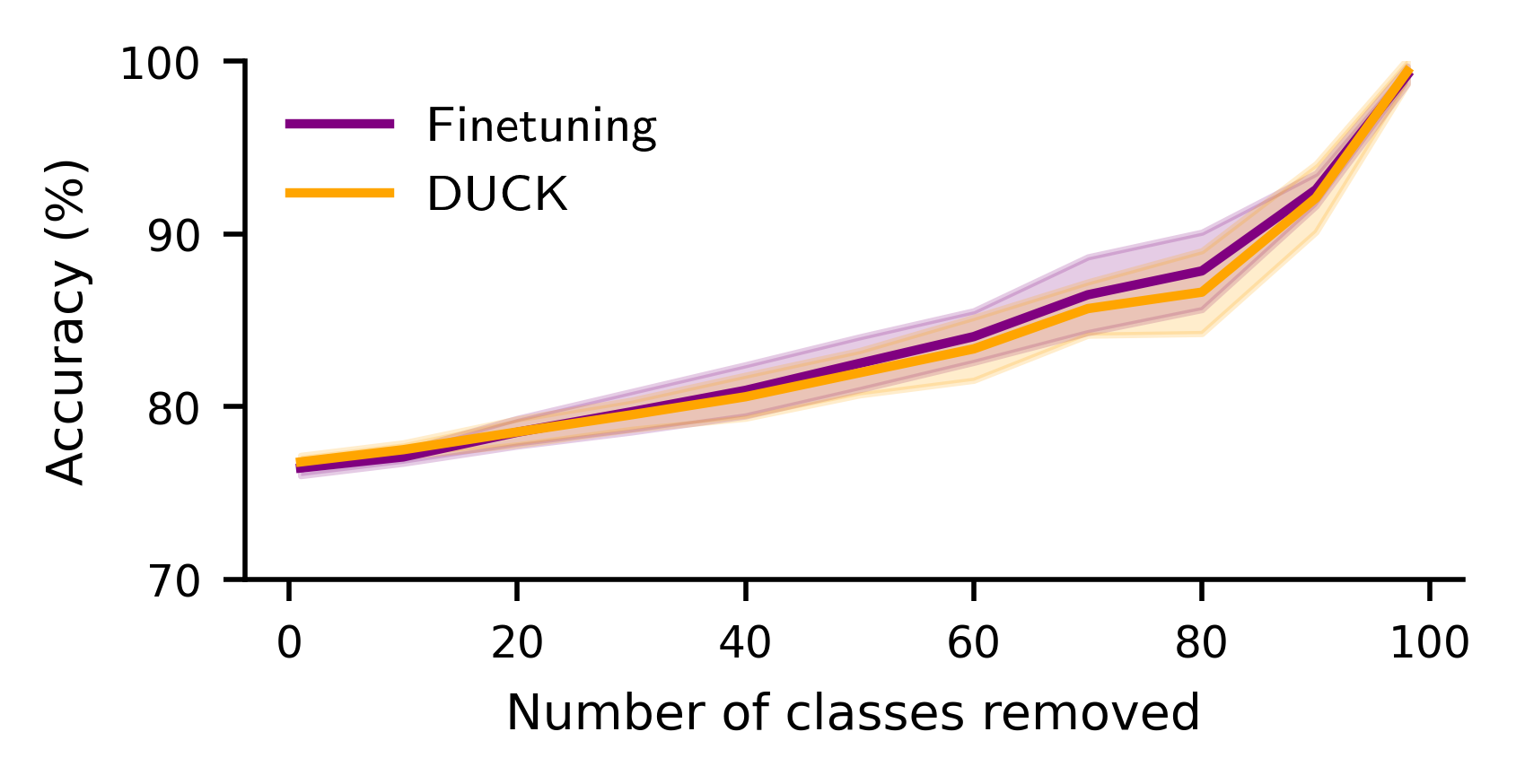}
    \caption{DUCK accuracy performance on the $\mathcal{D}_r^t$ of CIFAR100 (orange) compared to the model finetuned on $\mathcal{D}_r$ (purple) as a function of the number of classes removed. The accuracy and std reported are computed over 10 different shuffles of the classes.}
    \label{fig:multiple}
\end{figure}

In addition to CR and HR scenarios, we opted to assess the performance of our algorithm when multiple classes are removed from the dataset. This scenario represents a natural extension of CR, with more than a single class included in the forget-set. In Figure \ref{fig:multiple}, we present the results in terms of the retain test accuracy $\mathcal{A}^t_r$ as a function of the number of classes to be removed. The accuracy and std reported are computed over 10 different shufflings of the classes. $\mathcal{A}^t_f$ is compatible with 0 for both DUCK and Finetuning for any number of removed classes.
Remarkably, DUCK achieves compatible performance with the more resource-intensive Finetuning method even with a large number of classes in the forget-set. This result demonstrates how DUCK can remove much of the original model information without affecting its generalization capabilities.
\label{subsec:multiple}

\subsection{Unlearning Time Analysis}
\label{subsec:time}
\begin{figure}[t]
    \centering
    \includegraphics[width=.92\linewidth]{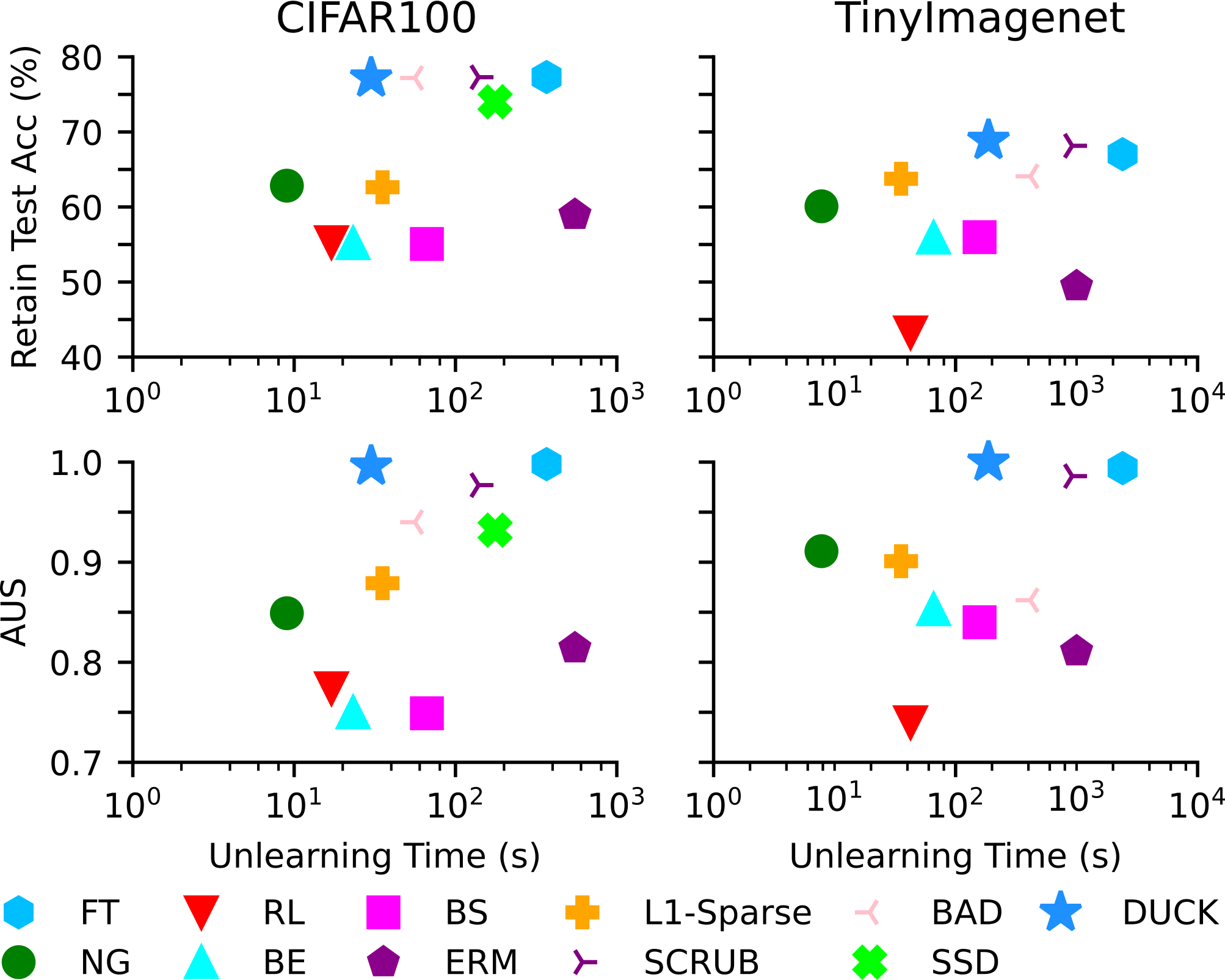}
    \caption{Average test accuracy over $\mathcal{D}^t_r$ and AUS as a function of the unlearning time in the CR scenario for CIFAR-100 and TinyImagenet. Accuracies and AUS are reported as mean across 10 runs where for each one a different target class was selected.}
    \label{fig:time}
\end{figure}

As a further analysis, we plot in Figure \ref{fig:time} the $\mathcal{A}_r^t$ and AUS as a function of the unlearning time for all the methods considered in the CR scenario. Notably, DUCK is consistently superior to SOTA unlearning methods for both values and is comparable to Finetuning, which is $\sim$10 times slower. Among the methods that do not deteriorate  $\mathcal{A}_r^t$ significantly, (DUCK, SSD, Bad-Teacher, and SCRUB) DUCK represents the most accurate and fastest alternative overall. From these results, it is possible to evince that DUCK is an effective method able to balance the time-accuracy tradeoff.
Furthermore, in Figure 1 of the Supp. Materials, we present a similar plot for the HR scenario, where the superiority of DUCK is even more pronounced.
\begin{table}
    \centering
    \caption{Results of the application of DUCK to different architectures trained on CIFAR100 and TinyImagenet in the CR scenario. For each model, we reported both the original model and the DUCK unlearned model performance. The accuracies reported here are the mean over 10 runs where for each one a different class has to be forgotten.}
\resizebox{1.\linewidth}{!}{\begin{tabular}{ccccccc}
\toprule
                                            &\multicolumn{3}{c}{Model}      & $\mathcal{A}^t_r$ & $\mathcal{A}^t_f$ & AUS \\
\midrule
\multirow{10}{*}{\rotatebox[]{90}{CIFAR100}}&\multirow{2}{*}{AllCNN}  
& \multirow{2}{*}{1.73M} & Or.  & 68.11 (00.17) & 68.10 (16.86) & 0.539 (0.054) \\
&&& DUCK & 67.57 (00.35) & 00.00 (00.00) & 0.995 (0.004) \\
\cmidrule{2-7}
&\multirow{2}{*}{Resnet18} 
& \multirow{2}{*}{11.23M} & Or.  & 77.55 (00.11) & 77.50 (02.80) & 0.563 (0.009) \\
&&& DUCK & 77.18 (00.28)    & 00.00 (00.00) & 0.996 (0.003) \\
\cmidrule{2-7}
&\multirow{2}{*}{Resnet34} 
& \multirow{2}{*}{21.34M} & Or.  & 80.14 (00.10) & 80.30 (10.02) & 0.555 (0.031) \\
&&& DUCK & 79.47 (00.25)    & 00.00 (00.00) & 0.993 (0.003) \\
\cmidrule{2-7}
&\multirow{2}{*}{Resnet50} 
& \multirow{2}{*}{23.71M} & Or.  & 80.64 (00.10) & 80.80 (09.54) & 0.553 (0.029) \\
&&& DUCK & 80.48 (00.27) & 00.00 (00.00) & 0.998 (0.003) \\
\cmidrule{2-7}
&\multirow{2}{*}{ViT}     
& \multirow{2}{*}{85.86M} & Or.  & 85.51 (00.11) & 83.40 (11.26) & 0.545 (0.033) \\
&&& DUCK & 85.71 (00.25) & 00.00 (00.00) & 1.002 (0.003) \\
\toprule
\multirow{10}{*}{\rotatebox[]{90}{TinyImagenet}}&\multirow{2}{*}{AllCNN}   
& \multirow{2}{*}{1.77M} & Or.  &52.79 (00.11) & 50.40 (16.07) & 0.665 (0.071) \\
&&& DUCK &  52.22 (00.16) & 00.00 (00.00) & 0.994 (0.002)\\
\cmidrule{2-7}
&\multirow{2}{*}{Resnet18} 
& \multirow{2}{*}{11.28M} & Or.  &68.26 (00.08) & 64.60 (15.64) & 0.607 (0.058) \\
&&& DUCK &68.47 (00.33) & 00.00 (00.00) & 1.002 (0.003)\\
\cmidrule{2-7}
&\multirow{2}{*}{Resnet34} 
& \multirow{2}{*}{21.39M} & Or.  &72.75 (00.07) & 71.40 (13.86) & 0.583 (0.047) \\
&&& DUCK &73.02 (00.18) & 00.00 (00.00) & 1.002 (0.002) \\
\cmidrule{2-7}
&\multirow{2}{*}{Resnet50} 
& \multirow{2}{*}{23.92M} & Or.  &75.36 (00.06) & 77.40 (11.93) & 0.564 (0.038)\\
&&& DUCK &75.46 (00.23) & 00.00 (00.00) & 1.001 (0.002) \\
\cmidrule{2-7}
&\multirow{2}{*}{ViT}      
& \multirow{2}{*}{85.95M} & Or.  & 84.98(00.05) & 83.40 (09.85) & 0.545 (0.029)  \\   
&&& DUCK & 85.00(00.24)    & 00.00 (00.00)     & 1.000 (0.002) \\
\bottomrule
\end{tabular}}

    \label{tab:arch}
\end{table}
\subsection{Architectural Analysis}
\label{subsubsec:architectural}
To assess the method's robustness against changes in the model architecture, we conducted experiments using various backbone architectures, specifically AllCNN, \texttt{resnet18}, \texttt{resnet34}, \texttt{resnet50} and ViT-b16 models. These models are listed in ascending order based on the number of parameters. The results regarding accuracy and AUS are detailed in Table~\ref{tab:arch}. Means and standard deviations are calculated over 10 runs in the CR scenario, varying the forget class. Observing the outcomes, we note that even if the original model $\mathcal{A}^t_r$ increases due to a large number of parameters, the AUS remains stable. These results highlight how DUCK regardless of the size and type of model can perform efficiently the unlearning task ($\mathcal{A}^t_f = 0$) without affecting the original model accuracy. Overall, DUCK represents a valid model-agnostic efficient solution to machine unlearning problems. 

\begin{figure*}
    \centering
    \includegraphics[width=.9\linewidth]{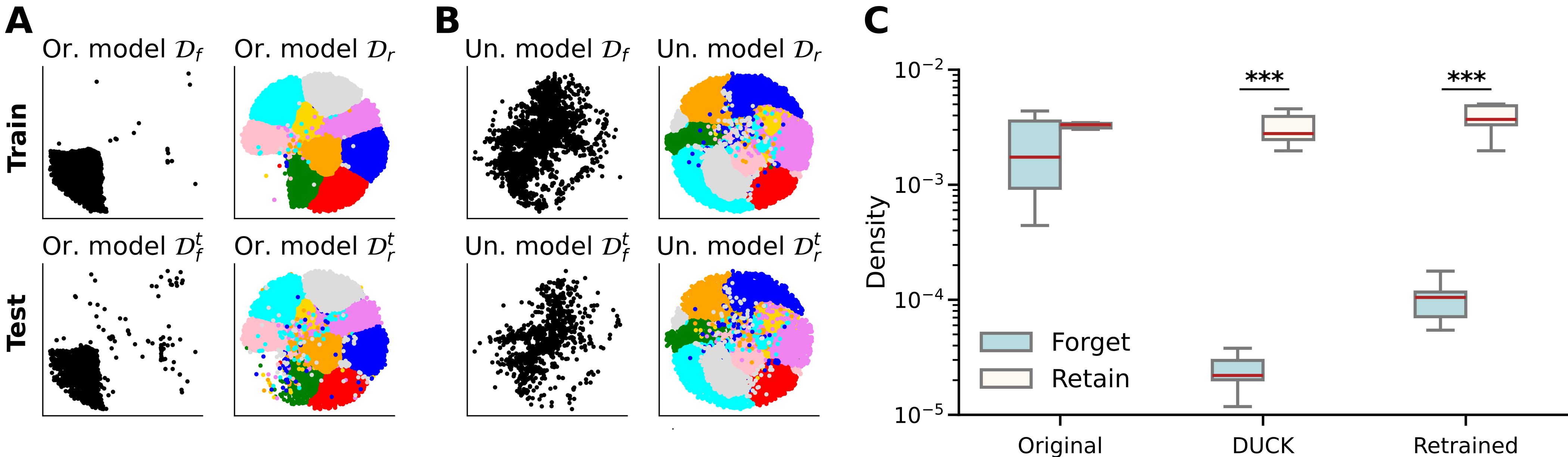}
    \caption{Analysis of the structure of the feature space. A-B) Visualization of $\mathcal{D}_r$, $\mathcal{D}_f$, $\mathcal{D}_r^t$ and $\mathcal{D}_f^t$ samples embeddings of CIFAR-10 dataset obtained using TSNE\cite{van2008visualizing} for CR scenario. Embeddings are represented for the original model (Or. model; A) and the unlearned model (Un. model; B). The first row is related to the train samples, and the second one to the test ones. Black dots represent the forget class that has to be removed (class 0 - plane, in this case). The colored dots represent samples belonging to different retain classes. Ideally, forget samples are scattered between all the remaining classes. C) Comparative analysis of feature cluster densities across different model states. Computation is performed across 10 different runs, each one with a different forget class. Median values are shown in red, the boxes show the quartiles of the dataset while the whiskers extend to show the rest of the distribution. Forget (Retain) clusters are represented in light green (white) for the Original, DUCK, and Retrained models. The y-axis represents the density on a logarithmic scale. * p$<$0.05, ** p$<$ 0.001 and ***  p$<$ 0.0001, 2-sided Wilcoxon signed-rank test}
    \label{fig:tsne}
\end{figure*}

\subsection{Analysis of the structure of the feature space}
\label{subsec:manifold}
To better understand the effect of the unlearning mechanism of our method, we analyzed the embeddings extracted from the backbone of the neural network ($\Psi_{\theta}$) both before and after the unlearning process in the CR scenario. In Figure~\ref{fig:tsne}-A/B, we report the application of t-SNE \cite{van2008visualizing} to the extracted embeddings, for the CIFAR10 dataset, with each class represented by a distinct color cluster. We fit 2 different t-SNEs respectively for the original model and the unlearned model embeddings. Before applying DUCK (Original model, Fig.~\ref{fig:tsne}A), embeddings corresponding to different classes are clustered together in both the train and test sets. After applying DUCK (Fig.~\ref{fig:tsne}B), while the embeddings of the retain-set samples are still clustered together, the embeddings of the forget-set are scattered across all the other class clusters. The results suggest how the information content of the embeddings of the forget-set extracted from $\Psi_{\theta}$ is no longer reliable and useful for the classification of these samples.

To quantitatively complement our t-SNE visualization, we further investigated the density of feature clusters in the high-dimensional space before the fully connected layer. After employing Principal Component Analysis (PCA) \cite{PCA} to reduce the dimensionality of this space, we were able to compute the hypervolume of the n-sphere containing 95\% of the data for each cluster.
We defined cluster density as the ratio between the number of samples $N_{s}(r)$ with distance from centroid $<r$, and the respective n-dimensional hypersphere's volume $V_n(r)$, as shown in Equation \ref{eq:density}.
\begin{equation}
    V_n(r) = \frac{\pi^{n/2}}{\Gamma(\frac{n}{2}+1)}\cdot r^n , \hspace{1cm} d(r) = \frac{N_{s}(r)}{V_n(r)}.
    \label{eq:density}
\end{equation}
$\Gamma$ is the gamma function and $n$ represents the number of dimensions of the hyper-space which in our case is $n=9$ after PCA reduction ( we set this value for covering 95\% of the total variance of the features in the feature space).
Our analysis spanned three conditions: the original model, the model post-retraining, and the model after applying DUCK.

This investigation allowed us to observe the cluster density dynamics effectively. Figure \ref{fig:tsne}-C shows the results for CIFAR10 over 10 runs, with the forget class changing for each run. In the original model setup, both forget and retain class clusters exhibited similar densities. However, post-unlearning, a marked density reduction was evident for forget class clusters in both the retrained and DUCK models, highlighting the dilution of feature space presence for these classes. Interestingly, the comparison between retrained and DUCK methods revealed remarkably similar densities for the forget-class clusters, confirming the efficacy of DUCK in dispersing forget-class features while maintaining the integrity of retained classes. This pattern supports our visual findings from the t-SNE analysis, further validating DUCK's role in inducing selective forgetting without degrading overall model performance.

\subsection{Interpretability}
A critical aspect regarding DUCK is how the metric learning mechanism can remove forget samples related information and induce forgetting. To investigate deeper these aspects we applied the SHapley Additive exPlanations (SHAP) approach \cite{SHAP}, a powerful method for explaining the output of machine learning models, including DNNs. It provides a way to understand the impact of each feature on the model's predictions. The key idea behind SHAP is rooted in cooperative game theory, specifically the concept of Shapley values. SHAP assigns each feature an importance score based on its contribution to the model's output across all possible combinations of features. This enables a nuanced understanding of how individual features influence the model's predictions.

\begin{figure*}[ht]
    \centering
    \includegraphics[width=.86\linewidth]{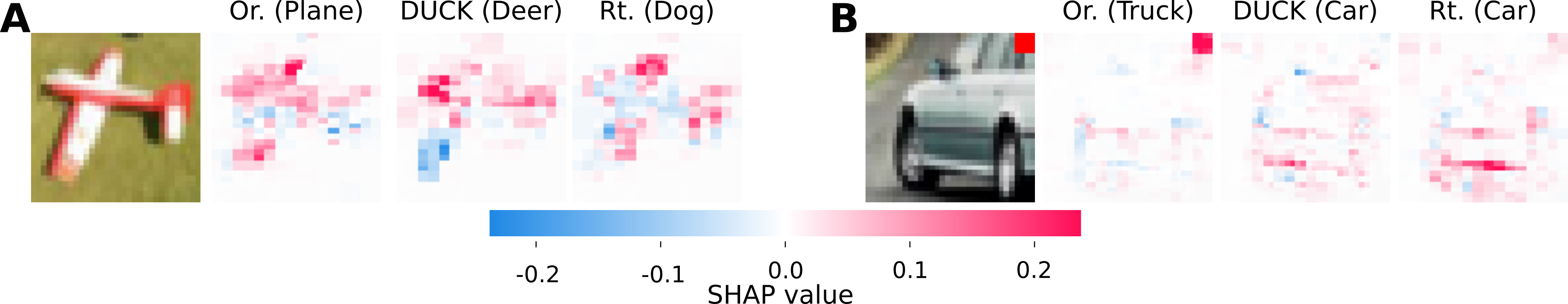}
    \caption{Visualization of the Shapley values, evaluating each feature's importance. Red areas of the image increase the probability of that class, and blue areas decrease the probability. \textbf{A}) Comparison among the output of, in order, Original model (Or.), DUCK, and Retrained model (Rt.) for a plane input image from the CIFAR10 dataset. \textbf{B}) Visual results of the bias removal experiment performed on the CIFAR10 dataset confounding car/truck images.}
    \label{fig:xai}
\end{figure*}
The SHAP method stands out as a powerful tool for delving deeper into the impact of feature importance during the unlearning process. As shown in Figure \ref{fig:xai}A, we utilize SHAP to assess how the original model, along with the model modified by DUCK and the retrained model, prioritize different features when analyzing an airplane image that belongs to the forget set. Notably, this comparison reveals that not only unlearned and retrained models make ``incorrect'' predictions, but also there are notable shifts in their feature importance maps relative to the original model. This finding suggests DUCK's effectiveness in diminishing class-specific information, leading to a transformative adjustment in how features are valued and utilized for samples designated to be forgotten.

To delve deeper into this phenomenon, we executed a bias removal experiment, depicted in Figure \ref{fig:xai}B. We initially trained a model (specifically, \texttt{resnet18}) on a version of the CIFAR10 dataset biased by the addition of a red 4x4 square to the top right corner of 200 car class samples, mislabeling them as trucks. To introduce further complexity, we inserted a confounder by applying the same red square to 200 truck class samples without changing their labels. Predictably, this introduced a strong bias, with the model incorrectly identifying images with the red square as trucks in a significant percentage of cases (99.8\%). By applying the DUCK method, we successfully corrected this bias, enabling the model to accurately classify mislabeled car images, achieving 98.0\% accuracy for unseen car+red square images. Afterward, for a comparison check, we trained again a model from scratch, correcting the labels of the biased car images, obtaining 97.8\% accuracy for unseen car+red square images (experimental details are reported in the Supp. Materials).

For each of these three models -original, unlearned by DUCK, and retrained-, we employed SHAP to calculate the mean SHAP value associated with the red square region. This analysis allowed us to measure how DUCK effectively diminishes the significance of the bias-inducing feature compared to the original and retrained models. Remarkably, the DUCK model exhibited a mean SHAP value (0.006$\pm$0.004) similar to that of the retrained model (0.007$\pm$0.004, $p=0.33$ 2-sided Wilcoxon signed-rank test), both of which were substantially reduced compared to the original biased model (0.06$\pm$0.01, $p<0.001$ for either unlearned or retrained vs. original 2-sided Wilcoxon signed-rank test). These results demonstrate the efficacy of DUCK in eliminating information specific to the biased examples, thereby restoring accurate model predictions. 

Thus, the metric learning mechanism embedded in DUCK emerges as a pivotal strategy for eradicating forgettable sample-related information, thereby modifying the feature extraction within the DNN and subsequently enhancing the overall DNN predictions.

\section{Conclusions}
\label{sec:conclusions}
In this study, we presented DUCK, a novel machine unlearning algorithm leveraging metric learning to precisely remove unwanted data from neural networks, thereby addressing the increasing need for privacy-preserving AI systems. Our experimental results showcased DUCK's effectiveness across several datasets and unlearning scenarios. DUCK not only outperforms all the tested methods from SOTA in terms of efficiency and efficacy but also achieves performances comparable with the retrained model. DUCK meets the growing demand for privacy-compliant AI systems but it also sets a new benchmark for the selective erasure of data from neural networks. The introduction of the Adaptive Unlearning Score (AUS) offers a new metric for evaluating unlearning performance, balancing data erasure against the preservation of model accuracy. Our qualitative and quantitative evaluations, particularly our exploration into the high-dimensional feature space, underscore the thoroughness of our approach in assessing the practical effectiveness of information erasure. The bias removal experiment specifically illustrates DUCK's applicability in addressing real-world challenges, showcasing its potential to correct biases within AI models effectively. This aspect of our research highlights the depth of our investigation into not only the algorithm's performance but also its implications for creating fairer and more ethical AI systems.

Future directions could explore the scalability of DUCK to larger, more complex datasets and its integration within different tasks such as natural language processing and reinforcement learning. Additionally, there's a promising avenue in enhancing DUCK's efficiency to support dynamic unlearning in real-time applications, further broadening its applicability and impact. Addressing these challenges will not only solidify the foundations of machine unlearning but also pave the way for AI systems that can adapt, forget, and relearn with minimal human intervention and maximum respect for privacy and ethical standards.

\section*{Supplementary Material}
\subsection{Datasets Details}
This section provides detailed information about the three publicly available datasets utilized in the experiments outlined in the main paper: CIFAR10\cite{krizhevsky2009learning}, CIFAR100\cite{krizhevsky2009learning}, and  TinyImagenet\cite{Le2015TinyIV}.

\minisection{CIFAR10 and CIFAR100.}
Both datasets consist of 60,000 images sized $32 \times 32$, split into a training set of 50,000 images and a test set of 10,000 images. CIFAR10 contains 10 classes, with 6,000 samples for each class, while CIFAR100 includes 100 classes, resulting in 600 images per class.

\minisection{TinyImagenet.}
Resized from the larger ImageNet dataset\cite{deng2009imagenet}, TinyImagenet comprises 110,000 images sized $64 \times 64$, distributed across 200 classes. The training set contains 100,000 images (500 per class), and the test set contains 10,000 images (50 per class), following a structure similar to CIFAR100.

\subsection{Experimental Details}
In this Section, we report the experimental details for the training and evaluation of DUCK and all the considered methods.

\minisection{DUCK Hyperparameters and setting.}
In Table \ref{tab:hyper_duck}, we present the hyperparameters employed in DUCK for all the considered scenarios. The upper section of the table pertains to the class-removal scenario (CR), while the lower section refers to the Homogeneous-Removal (HR) scenario. The Temperature parameter is utilized to scale probabilities in the computation of the cross-entropy loss (eq.  4).

For the multiple class removal experiment, which represents a challenging variant of the CR scenario, we maintained fixed hyperparameters while reducing the batch ratio to 1 when at least 50 classes are removed.
\begin{table}[!h]
\caption{Hyperparameters employed in DUCK for both the CR and HR scenarios.}
\label{tab:hyper_duck}
\begin{tabular}{ccccc}
\toprule
\multicolumn{1}{l}{}       && CIFAR10       & CIFAR100      & TinyImagenet    \\
\midrule

\multirow{5}{*}{\rotatebox[]{90}{CR}}
&$\lambda_{\text{fgt}}$     &   1.5         &   1.5         &   0.5          \\
&$\lambda_{\text{ret}}$     &   1.5         &   1.5         &   1.5          \\
&Batch Ratio                &   5           &   5           &   30           \\
&Learning Rate              &   \num{1e-3}  &   \num{1e-3}  &   \num{2e-4}   \\
&Batch Size                 &   1024        &   1024        &   1024         \\
&Temperature                &   2           &   2           &   3            \\

\midrule

\multirow{5}{*}{\rotatebox[]{90}{HR}}
&$\lambda_{\text{fgt}}$     &   1           &   1           &   1.5        \\
&$\lambda_{\text{ret}}$     &   1.4         &   1.4         &   1.5        \\
&Batch Ratio                &   5           &   5           &   5          \\
&Learning Rate              &   \num{1e-3}  &   \num{1e-3}  &   \num{5e-4} \\
&Batch Size                 &   1024        &   1024        &   1024       \\
&Temperature                &   2           &   2           &   3          \\
\bottomrule
\end{tabular}
\end{table}

\minisection{Baselines and competitors hyperparameters and settings.}
In Table \ref{tab:base_hyper_CR} and Table \ref{tab:base_hyper_HR}, we outline the hyperparameters utilized in conducting experiments with the baselines (Original, Retrained, and FineTuning) and the considered competitor methods (Random Labels, Negative Gradient, Boundary Shrink, Boundary Expanding, SCRUB, L1-Sparse, ERM-KTP, and SSD).

As specified in the main paper, we solely executed the HR scenario for the methods capable of supporting this setting. Notably, for Random Labels and Negative Gradient, the number of epochs is denoted as $\star$. This signifies that, in these experiments, they were executed for an unspecified number of epochs, continuing until they met the same stopping criteria as DUCK ($A^{\star}_f=0.01$ for CR and $A^{\star}_f=A^{\text{Or}}_t$ for HR). The symbol $-$ indicates that the learning rate scheduler was not utilized. For SSD\cite{foster2023SSD} we used $\alpha=10$ and $\lambda=1$ as specified in the corresponding paper.

\begin{table}[]
\caption{Hyperparameters utilized for experiments with baselines (Original, Retrained, and FineTuning) and the considered competitors in the class-removal (CR) scenario. For Random Labels and Negative Gradient, the number of epochs denoted by $\star$ indicates the experiments ran until meeting the stopping criteria akin to DUCK ($A^{\star}_f=0.01$ for CR). The symbol $-$ signifies that the learning rate scheduler was not employed.}
\label{tab:base_hyper_CR}

\resizebox{1.\linewidth}{!}{\begin{tabular}{cccccc}

\toprule
    &                    & Learning Rate & Batch Size & Epochs  & Scheduler\\
    \midrule
\multirow{12}{*}{\rotatebox[]{90}{CIFAR10}}
    & Original           & \num{1e-1}    & 256        & 200     &\text{Cosine-Annealing}\\
    & Retrained          & \num{1e-1}    & 256        & 200     &\text{Cosine-Annealing}\\
    & FineTuning         & \num{1e-1}    & 32         & 30      &[8,15]\\
    & Negative Gradient  & \num{1e-2}    & 256        & $\star$ &-\\
    & Random Labels      & \num{1e-3}    & 256        & $\star$ &[3]\\
    & Boundary S.        & \num{1e-5}    & 64         & 15      &-\\
    & Boundary E.        & \num{1e-5}    & 64         & 10      &-\\
    & SCRUB              & \num{5e-4}    & 256        & 3 (2)   &[3]\\
    & L1-Sparse          &\num{1e-2}     &256         &10       &-\\
    & ERM-KTP            & \num{1e-1}    & 256        & 50      &[10,20] $\gamma=0.1$\\
    & Bad Teacher        & \num{5e-4}    &256&3&-\\

\midrule

    &                    & Learning Rate & Batch Size & Epochs  & Scheduler\\
    \midrule
\multirow{12}{*}{\rotatebox[]{90}{CIFAR100}}
    & Original           & \num{1e-1}    & 256        & 200     &\text{Cosine-Annealing}\\
    & Retrained          & \num{1e-1}    & 256        & 200     &\text{Cosine-Annealing}\\
    & FineTuning         & \num{1e-1}    & 32         & 30      &[8,15]\\
    & Negative Gradient  & \num{1e-2}    & 256        & $\star$ &-\\
    & Random Labels      & \num{1e-2}    & 256        & $\star$ &-\\
    & Boundary S.        & \num{1e-5}    & 64         & 15      &-\\
    & Boundary E.        & \num{1e-5}    & 64         & 10      &-\\
    & SCRUB              & \num{5e-3}    &256         & 10(8)   &[9]\\
    & L1-Sparse          &\num{1e-2} &256&10&-\\
    & ERM-KTP            & \num{1e-1}    & 256        & 50      &[10,20] $\gamma=0.1$\\
    & Bad Teacher        & \num{5e-4}    & 256        & 3       &-\\
    & SSD                &\num{1e-1}          & 64         &-        &-\\
    \midrule

    &                    & Learning Rate & Batch Size & Epochs  & Scheduler\\
    \midrule
\multirow{12}{*}{\rotatebox[]{90}{\begin{tabular}[c]{@{}c@{}}Tiny\\ Imagenet\end{tabular}}}
    & Original           & \num{1e-1}    & 256        & 200     & [every 25] $\gamma=0.1$\\
    & Retrained          & \num{1e-1}    & 256        & 200     & [every 25] $\gamma=0.1$\\
    & FineTuning         & \num{1e-1}    & 32         & 30      &[8,15]\\
    & Negative Gradient  & \num{1e-2}    & 256        & $\star$ &-\\
    & Random Labels      & \num{2.5e-3}  & 256        & $\star$ &-\\
    & Boundary S.        & \num{1e-5}    & 64         & 15      &-\\
    & Boundary E. & \num{1e-5}    & 64 &10&-\\
    & SCRUB              &\num{5e-4}     & 256        & 10 (8)  &[9]\\
    & L1-Sparse          &\num{1e-3} &256&10&-\\
    & ERM-KTP            & \num{1e-1}    & 256        & 50      &[10,20] $\gamma=0.1$\\
    & Bad Teacher        &\num{1e-4}&256&5&-\\
       
    \bottomrule
\end{tabular}}

\end{table}

\begin{table}[]
\caption{Hyperparameters used for experiments with baselines (Original, Retrained, and FineTuning) and supported competitors in the Homogeneous-Removal (HR) scenario. The HR experiments were conducted for methods compatible with this setting. For Random Labels and Negative Gradient, the $\star$ symbol implies that the experiments were executed until reaching the same stopping criteria as DUCK ($A^{\star}_f=A^{\text{Or}}_t$ for HR). The presence of $-$ denotes that the learning rate scheduler was not employed.}
\label{tab:base_hyper_HR}
\sisetup{mode = text, text-font-command = \tiny}
\tiny

\resizebox{1.\linewidth}{!}{\begin{tabular}{cccccc}

\toprule
    &                    & Learning Rate & Batch Size              & Epochs       & Scheduler\\
    \midrule
\multirow{9}{*}{\rotatebox[]{90}{CIFAR10}}
    & Original           & \num{1e-1}    & 256 & 200&\text{Cosine-Annealing}\\
    & Retrained          & \num{1e-1}    & 256 & 200 &\text{Cosine-Annealing}\\
    & FineTuning         & \num{1e-1}    & 32  & 30 &[8,15]\\
    & Random Labels      & \num{1e-4}    & 256 & $\star$ &[5]\\
    & Negative Gradient  & \num{3e-4}    & 256 & $\star$ &[4]\\
    & SCRUB              & \num{5e-4}    & 256        & 3 (2)   &[3]\\
    & L1-Sparse          &\num{1e-2} &256&10&-\\
    & Bad Teacher        &\num{5e-4}&256&3&-\\    

\midrule

    &                    & Learning Rate & Batch Size              & Epochs       & Scheduler\\
    \midrule
\multirow{9}{*}{\rotatebox[]{90}{CIFAR100}}
    & Original           & \num{1e-1}    & 256 & 200 &\text{Cosine-Annealing}\\
    & Retrained          & \num{1e-1}    & 256 & 200 &\text{Cosine-Annealing}\\
    & FineTuning         & \num{1e-1}    & 32  & 30 &[8,15]\\
    & Random Labels      & \num{1.2e-3}  & 256 &$\star$&[9]\\
    & Negative Gradient  & \num{4e-4}    & 256 &$\star$&[7,9]\\
    & SCRUB              &\num{5e-3}     &256 &15(15)&[5,10]$\gamma=0.5$\\
    & L1-Sparse          &\num{1e-2} &256&10&-\\
    & Bad Teacher        &\num{5e-4}&256&3&-\\
    & SSD                &\num{1e-1}            & 64         &-        &-\\
    \midrule

    &                    & Learning Rate           & Batch Size              & Epochs       & Scheduler\\
    \midrule
\multirow{9}{*}{\rotatebox[]{90}{\begin{tabular}[c]{@{}c@{}}Tiny\\ Imagenet\end{tabular}}}
    & Original           & \num{1e-1}    & 256 & 200 & [every 25] $\gamma=0.1$\\
    & Retrained          & \num{1e-1}    & 256 & 200 & [every 25] $\gamma=0.1$\\
    & FineTuning         & \num{1e-1}    & 32  & 30 & [8,15]\\
    & Random Labels      & \num{1e-3}    & 256 & $\star$&[7]\\
    & Negative Gradient  & \num{6e-5}    & 256 &$\star$&[5,12]\\
    & SCRUB              &\num{2e-3}     & 256        & 5 (5)  &[1,3]\\
    & L1-Sparse          &\num{1e-3} &256&10&-\\
    & Bad Teacher        &\num{1e-4}&256&5&-\\
       
    \bottomrule
\end{tabular}}

\end{table}

\minisection{Code Execution Details.} In this section we report the experimental details for reproducing the experiments of the paper.
We first trained the Original models for the four considered datasets using the seed 42.

\noindent For the \textbf{CR} scenario, for all the datasets considered, and all the unlearning methods, the experimental steps were:
\begin{enumerate}
    \item Fix the seed to 42.
    \item Split the dataset in retain (train/test) and forget (train/test) composed by all the instances of a class $i$. 
    \item Run the unlearning.
    \item Evaluate the unlearned model on the retain (train/test) and forget (train/test) sets.
    \item Repeat stpes (2)-(4) selecting a different class $i$.
    \item Compute metrics averaging on the indices $i$ representing the classes removed at each iteration. 
\end{enumerate}
In this scenario, for CIFAR10, each iteration utilized one of the ten classes as the forget set. For CIFAR100 and TinyImagenet, the forget-set included classes that were multiples of 10 and 20 starting from 0 (e.g., CIFAR100: 0, 10, 20, \dots, 90; TinyImagenet: 0, 20, 40, \dots, 180).

\noindent For the \textbf{HR} scenario, for all the datasets considered, and all the methods, the experimental steps were:
\begin{enumerate}
    \item Fix the seed $i$
    \item Generate the forget set randomly picking 10\% of the images from the original training dataset. Generate the retain-set with the remaining 90\% 
    \item Run the unlearning.
    \item Evaluate the unlearned model on the retain, forget, and test sets.
    \item Repeat stpes (1)-(4) selecting a different seed $i$.
    \item Compute metrics averaging on the indices $i$ representing the seeds. 
\end{enumerate}
The experiments were conducted using the following seeds: $\left[0,1,2,3,4,5,6,7,8,42 \right]$. 

These experimental protocols were consistently applied across all experiments presented in the paper, encompassing ablation studies, multiple class removal scenario, time analysis, and architectural analysis.

All the experiments in the paper have been conducted using a workstation equipped with an Intel Xeon W-225 CPU @ 4.10GHz, 256Gb of RAM, and two NVIDIA RTX A6000 with 48Gb of memory (the experiments have been conducted using a single GPU and multi-gpu scenarios have been avoided).

\subsection{Additional Time Analysis}
\begin{figure}[h]
    \centering
    \includegraphics[width=\linewidth]{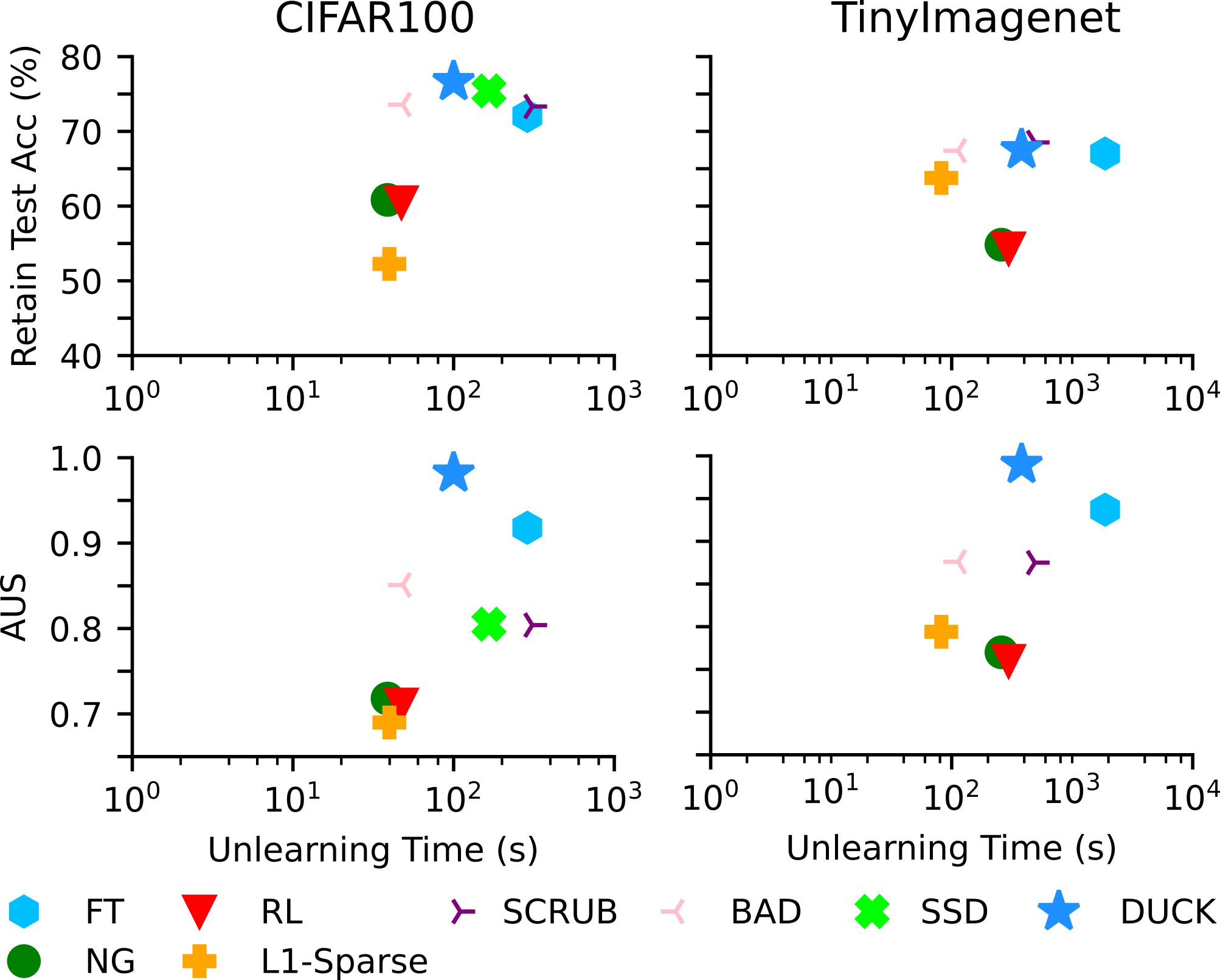}
    \caption{Unlearning time comparison in the HR scenario. Average test accuracy $\mathcal{D}^t$ and AUS as a function of the unlearning time for CIFAR-100 and TinyImagenet. Accuracies and AUS are reported as mean across 10 runs where for each one a different seed and forget split was selected.}
    \label{fig:time_supp}
\end{figure}

In Figure \ref{fig:time_supp} we report the test accuracy and AUS score as a function of the unlearning time in the HR scenario for CIFAR100 and TinyImagenet dataset. The results reported confirm once more how DUCK outperforms all other state-of-the-art methods and is faster and more accurate in terms of AUS than Finetuning.

\subsection{Additional details on Membership Inference Attacks}
\subsection{Offline LIRA}
\begin{figure}[!ht]
    \centering
    \includegraphics[width=\linewidth]{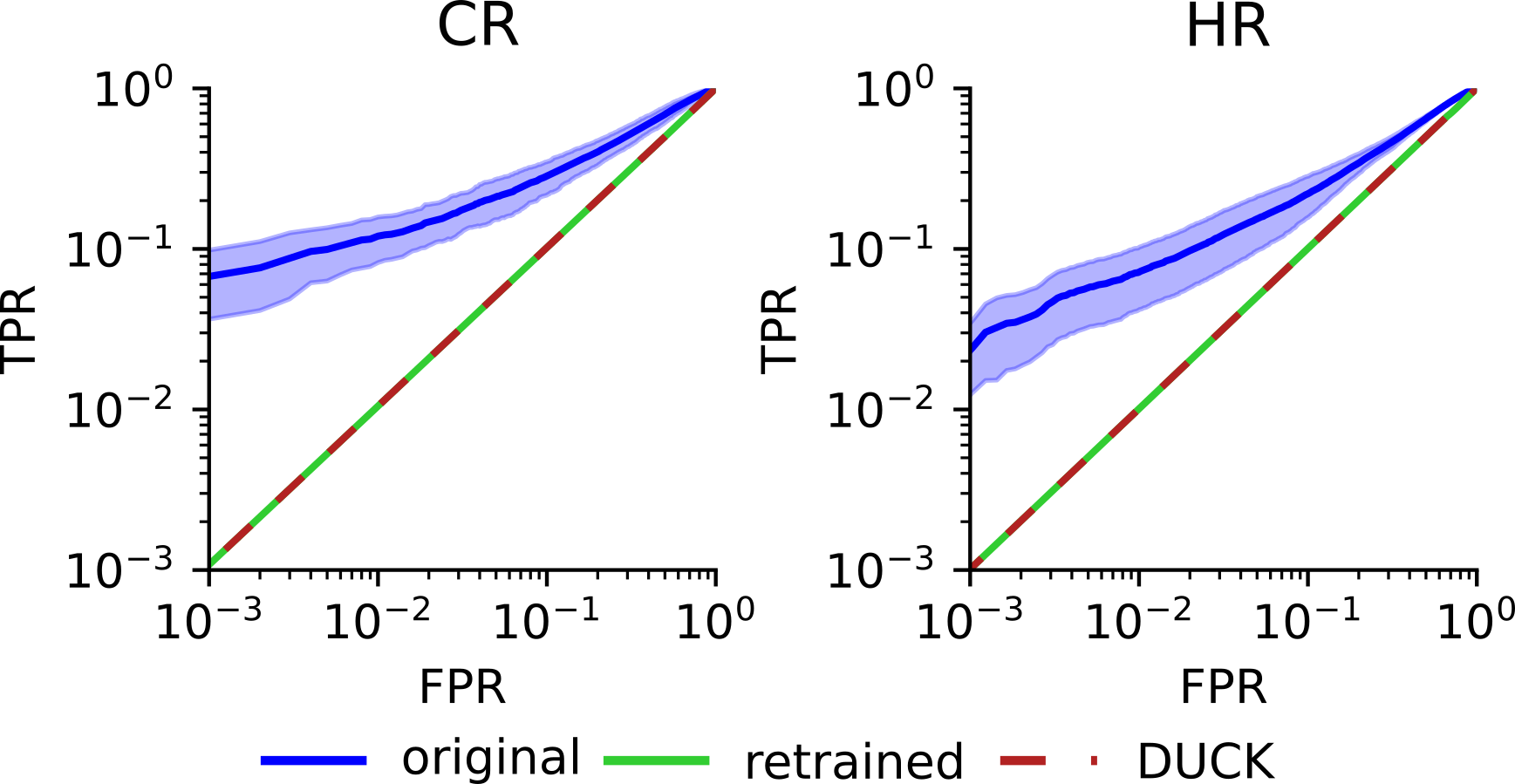}
    \caption{Receiver operating characteristic curve of the Original, unlearned by DUCK and retrained models in CIFAR10 for CR and HR scenarios. Curves are reported as mean $\pm$ std across 5 forget classes=0,1,2,3,4 for CR and for 5 seeds for HR}
    \label{fig:LIRA_roc}
\end{figure}

We reported the mean Receiver operating characteristic curve of the Original, unlearned by DUCK, and retrained models in CIFAR10 for CR and HR scenarios (Figure \ref{fig:LIRA_roc}). From these curves, we computed the True Positive Rate TPR at different low False Positive Rate FPR and the corresponding AUC reported in the main paper.

\begin{figure*}[!ht]
    \centering
    \includegraphics[width=\linewidth]{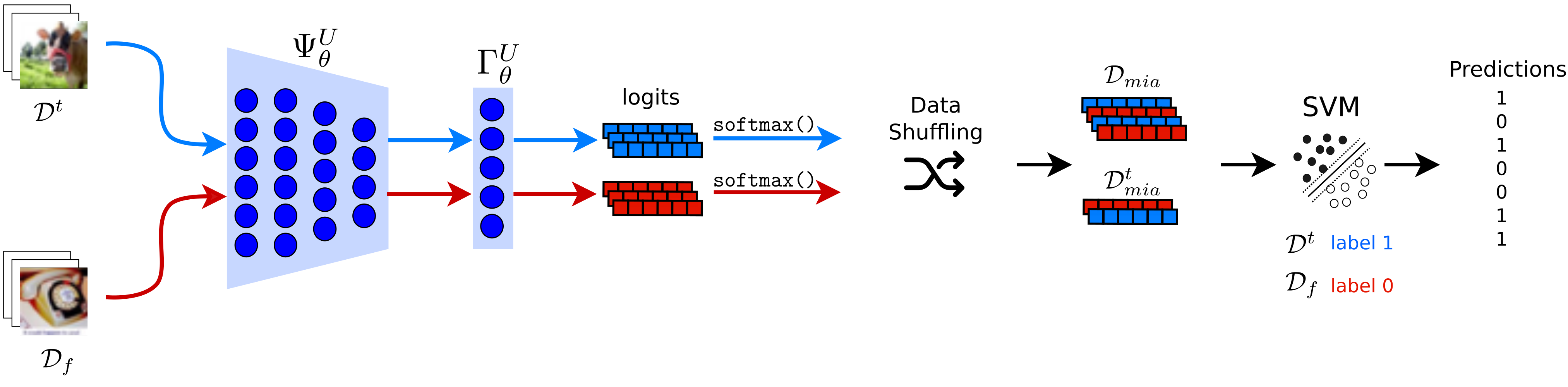}
    \caption{Architecture scheme representing the Single Classifier Membership Inference Attack (MIA). The logits vectors of datasets $\mathcal{D}_f$ (forget-set) and $\mathcal{D}^t$ (test-set) are combined, shuffled, and split into a training set $\mathcal{D}_{\text{mia}}$ (80\% of samples), and a test set $\mathcal{D}_{\text{mia}}^{t}$ (20\% of samples).  An SVM with a Gaussian kernel classifier is then trained to distinguish between $\mathcal{D}_f$ and $\mathcal{D}^t$ instances.}
    \label{fig:mia}
\end{figure*}

\subsection{Single Classifier Membership Inference Attack} 
The offline LIRA used in the main paper represents a class of Membership Inference Attacks (MIA) based on the usage of a large set of shadow models. This technique generates an empirical null-hypothesis distribution by which the membership of each forget set sample can be verified. Nevertheless, there is a class of MIA based on a single classifier model \cite{Salem2019,song2020} that identifies whether input samples were utilized during the training of the target model $\Phi_{\theta}$. Importantly, this single classifier MIA has the advantage of being extremely less computationally expensive at the cost of losing predictive power on the data membership \cite{Carlini22}. This single classifier MIA merges $\mathcal{D}_f$ and $\mathcal{D}^t$ and splits the resulting set into a training set $\mathcal{D}_{\text{mia}}$ (80$\%$ of samples) and a test set $\mathcal{D}_{\text{mia}}^{t}$ (20$\%$ of samples). It then builds an SVM with a Gaussian kernel classifier. The SVM takes as input the probability distribution across classes obtained from $\texttt{softmax}[\Phi^U_{\theta}(x_i)]$ where $x_i \in \mathcal{D}_{\text{mia}}$ or $\mathcal{D}_{\text{mia}}^{t}$. The SVM is trained on $\mathcal{D}_{\text{mia}}$ to separate training from test instances. The hyperparameters are tuned through a 3-fold cross-validation grid search and tested on $\mathcal{D}_{\text{mia}}^{t}$. The varied hyperparameters include the regularization parameter $C$, which takes values in the set $\left[1, 5, 10, 100\right]$, and the kernel coefficient $\gamma$, which takes values in the set $\left[1, 0.1, 0.01\right]$.
To better clarify the working mechanism of this MIA, we report the scheme in Figure \ref{fig:mia}. Failure of the MIA points out that the information about forget-samples has been removed from the unlearned model. The MIA performance is measured in terms of the mean F1-score over 5 runs of the SVM, each with different train and test splits of $\mathcal{D}_f$ and $\mathcal{D}^t$. The optimal F1 score is the one obtained with the retrained model. 

In the class-removal (CR) scenario, the forget-set comprises solely samples from a single class, unlike the heterogeneous removal (HR) scenario, where the test set consists of instances from multiple classes. Applying Membership Inference Attack (MIA) as previously defined could lead the SVM to detect biases in the logits associated with class identity, thus distorting results away from reflecting true sample membership. To counteract the introduction of such biases resulting from differing class distributions between the two datasets, we utilized the forget subset of the test data to ensure a fair and unbiased comparison. Instead of employing the entire test dataset $\mathcal{D}^t$ for MIA testing, we combined $\mathcal{D}_f$ with $\mathcal{D}_f^t$ to ensure consistent class identification across both subsets, thereby mitigating the introduction of class-related biases in $\mathcal{D}^t$. The SVM is trained to differentiate between forget train instances and forget test instances. Nevertheless, the number of forget samples ($N=5000$ for CIFAR10 and $N=500$ for CIFAR100) and forget test samples ($N=1000$ for CIFAR10 and $N=100$ for CIFAR100) are unevenly distributed. To address this imbalance, we resample $\mathcal{D}f$ to create a less imbalanced $\mathcal{D}{\text{mia}}$ with a ratio of 1:3 forget test samples per forget samples, resulting in a chance level of $.75$. Results for the CR scenario in CIFAR10 and CIFAR100 are presented in Table \ref{tab:mia_naive_table}.

In the HR scenario, both $\mathcal{D}^t$ and $\mathcal{D}_f$ contain all classes and an equal number of samples per dataset. Consequently, the chance level for the single classifier MIA in this case is $.5$.

Overall, despite the reduced predictive power of this MIA the results obtained confirm the results obtained with the offline LIRA presented in the main paper.
\begin{table}[]
\caption{Single Classifier MIA results for CR and HR in CIFAR10 and CIFAR100. Results are mean $\pm$ std across 10 runs where in CR a single class was removed or in HR a random subset of $\mathcal{D}$ was removed}
\label{tab:mia_naive_table}
\resizebox{1.\linewidth}{!}{\begin{tabular}{lcc|cc}
\toprule
                        & \multicolumn{2}{c}{CIFAR10}&\multicolumn{2}{c}{CIFAR100} \\
                  & CR  & HR &CR&HR \\
\midrule
Original      & 77.07 (00.49) & 54.62 (00.80)  &81.84 (04.22) & 61.80 (00.59)  \\
Retrained     & 74.68 (00.14) & 50.22 (00.80)  &76.31 (01.54) & 49.65 (00.40) \\
DUCK          & 74.72 (00.16) & 50.06 (00.64)  &76.89 (01.07) & 50.73 (00.56)  \\
\bottomrule
\end{tabular}}

\end{table}

\subsection{Bias Removal Experimental details}
In the bias removal experiment, we initially trained the original model for 200 epochs using a batch size of 256, a learning rate of 0.1, and Adam as the optimizer, employing Cosine Annealing as the scheduler. We utilized a biased version of CIFAR10, where 4x4 red squares were added to the top right corner of 200 images of cars and 200 images of trucks. Specifically, only for the 200 images of cars, the labels were altered to correspond to the class of trucks. The retrained model underwent training using the same optimization procedure and hyperparameters as the original model. In this process, the 200 mislabeled car images with the red square were corrected.

DUCK addressed the bias in the 200 images using the hyperparameters outlined in Table \ref{tab:hyper_duck}, as employed for CIFAR10 in the class-removal (CR) scenario. Unlike the CR and HR scenarios, the forget set consisted of the specific 200-biased images of cars. Notably, to evaluate whether our closest-centroid matching mechanism effectively removes bias, the embeddings of the forget set were matched with the centroid of the car class during the two-phase optimization procedure. Overall, no other modifications to the DUCK optimization procedure were made.
\newpage
\bibliographystyle{IEEEtran}
\bibliography{Manuscript}

\end{document}